\documentclass[10pt,twocolumn,letterpaper]{article}

\usepackage[pagenumbers]{cvpr}      %

\usepackage{graphicx}
\usepackage{amsmath}
\usepackage{amssymb}
\usepackage{booktabs}

\usepackage{pifont}
 \usepackage{array,multirow}

\usepackage{wrapfig}
\usepackage[pagebackref,breaklinks,colorlinks]{hyperref}

\usepackage[capitalize]{cleveref}
\crefname{section}{Sec.}{Secs.}
\Crefname{section}{Section}{Sections}
\Crefname{table}{Table}{Tables}
\crefname{table}{Tab.}{Tabs.}

\def\cvprPaperID{*****} %
\def\confName{CVPR}
\def\confYear{2022}

\begin{document}

\title{ SCR: Smooth Contour Regression with Geometric Priors}

\author{Gaétan Bahl\textsuperscript{1,2} \quad \quad Lionel Daniel\textsuperscript{1} \quad \quad Florent Lafarge\textsuperscript{2}\\
\textsuperscript{1}IRT Saint Exupéry \quad  \textsuperscript{2}Université Côte d'Azur - Inria \\
{\tt\small \{gaetan.bahl,florent.lafarge\}@inria.fr,  lionel.daniel@irt-saintexupery.com}
}

\maketitle

\newcommand{\loss}[1]{\mathcal{L}_\text{#1}}
\newcommand{\lambdaloss}[1]{\lambda_\text{#1}}
\newcommand{\Ntxt}[1]{N_\text{#1}}

\begin{abstract}

While object detection methods traditionally make use of pixel-level masks or bounding boxes, alternative representations such as polygons or active contours have recently emerged.  Among them, methods based on the regression of Fourier or Chebyshev coefficients have shown high potential on freeform objects. By defining object shapes as polar functions, they are however limited to star-shaped domains. We address this issue with  SCR: a method that captures resolution-free object contours as complex periodic functions. The method offers a good compromise between accuracy and compactness thanks to the design of efficient geometric shape priors. We benchmark  SCR on the popular COCO 2017 instance segmentation dataset, and show its competitiveness against existing algorithms in the field.  In addition, we design a compact version of our network, which we benchmark on embedded hardware with a wide range of power targets, achieving up to real-time performance.
\end{abstract}

\section{Introduction}

Over the recent years, the development of Convolutional Neural Networks \cite{Lecun1990}, combined
with the wide availability of powerful GPU hardware and large amounts of data for training \cite{JiaDeng2009},
has allowed great advances in various computer vision tasks, such as image classification \cite{krizhevsky2012imagenet},
object detection \cite{Girshick2014}, or instance segmentation \cite{he2017mask}.

Object detection is the task of finding bounding boxes of objects in images. 
However, an inherent flaw of bounding boxes is the fact that their IoU (Intersection over Union) with their 
actual underlying objects will always be limited by their rectangular shape. %
This is especially true for object that have holes or multiple sharp angles (\eg animals, bicycles).

Instance segmentation tries to alleviate this problem by providing pixel-level masks for each detected object.
Masks, however, 
also have some drawbacks as an object contour representation: they are resolution dependent, and have to be resized and interpolated to be used on images with different resolutions. Masks also take up a lot of storage space and require more computational power, which is especially problematic on embedded systems, such as drones or satellites. 
Furthermore, Geographic Information System (GIS) applications such as online mapping rely on vector formats, and the automatic conversion of raster masks to more compact representations such as polygons is a difficult scientific challenge \cite{li2020approximating}.

\newlength{\teaserheight}
\setlength{\teaserheight}{0.245\columnwidth}
\newlength{\teaserheightt}
\setlength{\teaserheightt}{0.284\columnwidth}
\newlength{\teaserheighttt}
\setlength{\teaserheighttt}{0.225\columnwidth}
\begin{figure}[t!]
    \centering
    \includegraphics[height=\teaserheight]{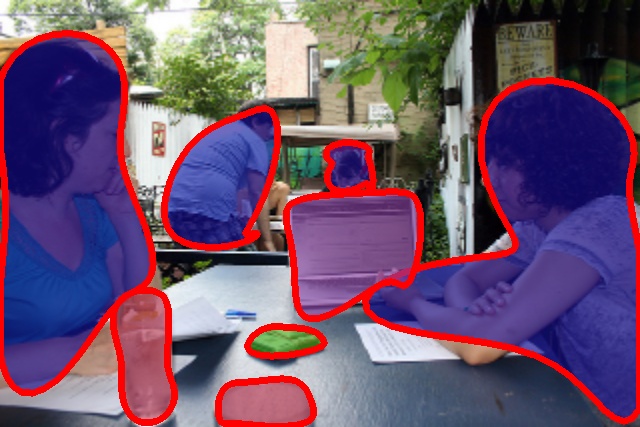}
    \includegraphics[height=\teaserheight]{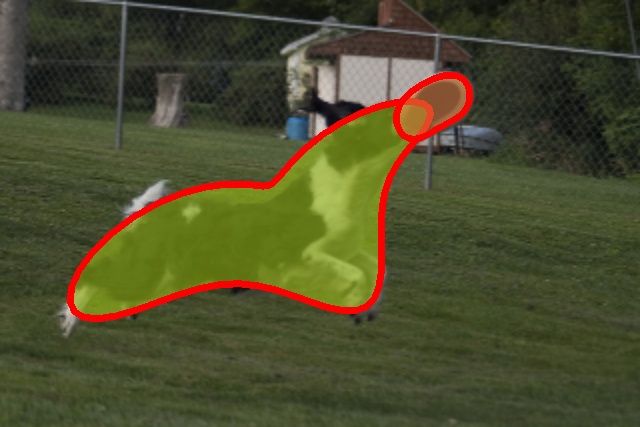}
    \includegraphics[height=\teaserheight]{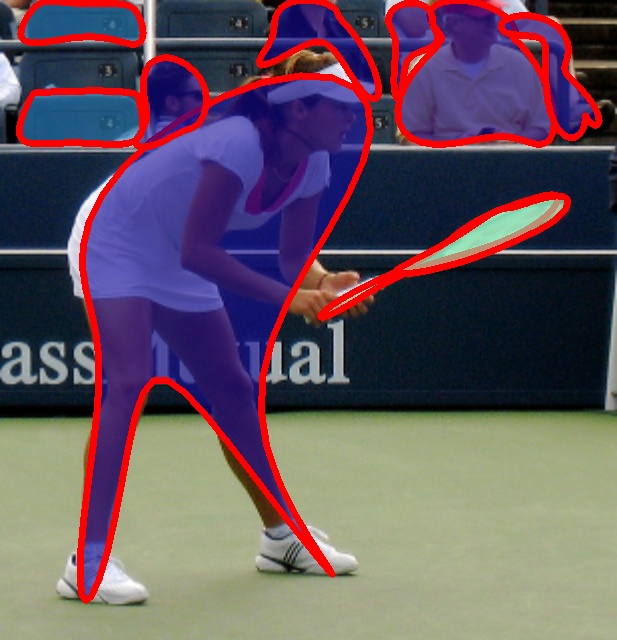}
    
    \includegraphics[height=\teaserheightt]{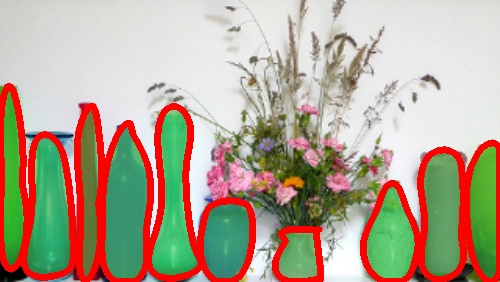}
    \includegraphics[height=\teaserheightt]{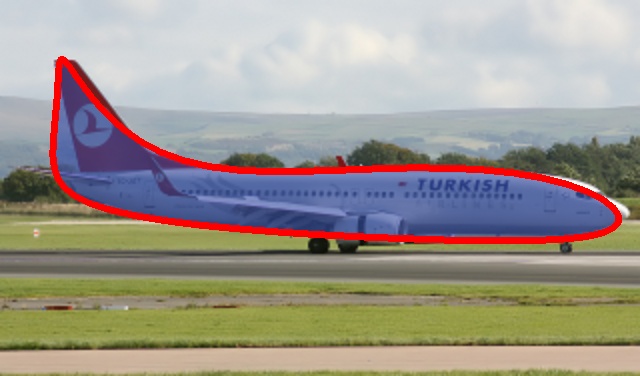}

    \includegraphics[height=\teaserheighttt]{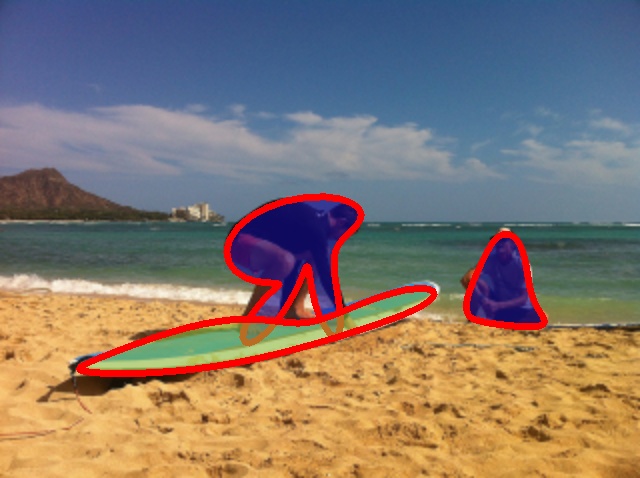}
    \includegraphics[height=\teaserheighttt]{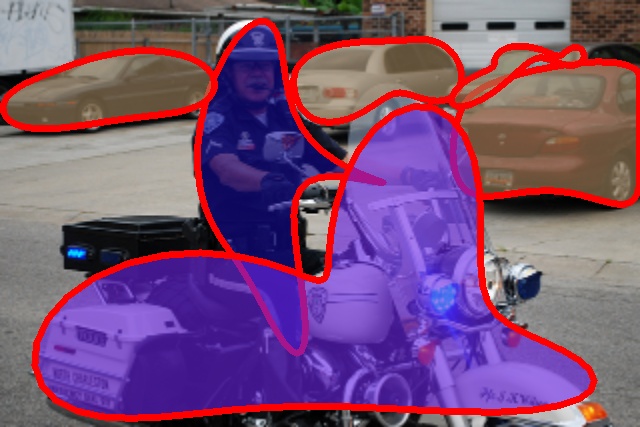}
    \includegraphics[height=\teaserheighttt]{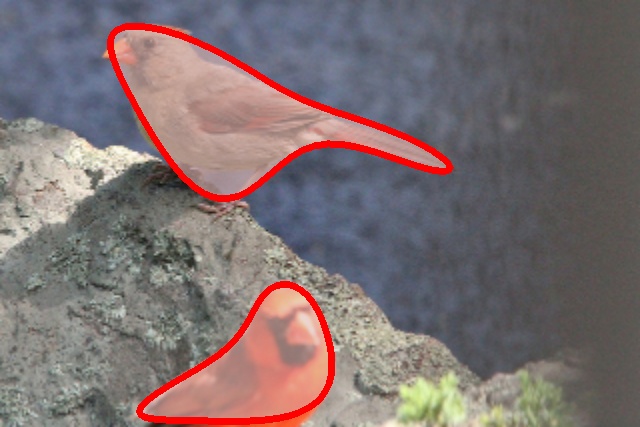}
    \caption{Our algorithm captures the silhouette of objects with a compact resolution-independent representation based on Fourier coefficients%
    . Contours produced by our algorithm adequately approximate the general shape of free-form objects such as persons, animals, cars or pots while being defined in a simple parametric way with only a few complex Fourier coefficients, here 8. Images from COCO 2017 test-dev.}%
    \label{fig:teaser}
\end{figure}

A good middle ground between bounding-box regression and mask-based instance segmentation is probably the regression of object contours using shape encoding, the goal being to capture the boundary of objects using a simple parametric function.
Methods such as PolarMask \cite{Xie2019} regress the contour of objects directly at a fixed set of sampling
points revolving around the object's center.
Some other methods use active contours or snake algorithms \cite{peng2020deep}.
Finally, methods such as ESE-Seg \cite{Xu2019} and FourierNet \cite{Benbarka2020}, make use of Chebyshev or Fourier coefficients to
explicitly regress an encoding that represents the shape of the object.

Our work proposes a Fourier shape decoder that brings several important improvements over previous methods. First, we give insights and mathematical arguments about the fact that Fourier coefficients constitute a better representation than the Chebyshev coefficients used in \cite{Xu2019}. By careful design of the loss function, we also improve the visual quality of regressed contours 
with fewer Fourier coefficients, %
while also simplifying the training process. Moving to a complex representation for contours allows us halve the number of calls to the Inverse Fast Fourier Transform (IFFT) in the final shape decoding stage, while allowing more freedom in shape representation by alleviating the limits of polar coordinates-based methods. Finally, we propose a compact version of our architecture based on a lightweight backbone \cite{Redmon2018} 
for use on low-power systems, typically found on board satellites and UAVs, 
targeting applications such as real time on-board instance segmentation and efficient transmission within
constellations of satellites.

\section{Related Work}

\paragraph{Detection neural networks}

R-CNN \cite{Girshick2014} and its follow-ups \cite{girshick2015fast, Ren2017}, 
were the original neural networks for object detection. They are two-stage approaches, where object locations
are first proposed by a Region Proposal Network (stage one) and then classified by a subsequent CNN (stage two).
Later, fully convolutional single-stage approaches were proposed \cite{Liu2015, Lin2017, Redmon2015}, 
which were significantly faster than their two-stage counterparts. These methods rely
on a set of "anchor" bounding boxes, to which the detected boxes are assigned and regressed
relatively. 
The simplicity and efficiency of these single stage detectors have made them
very popular over the years. A large number of improvements to this design have been
proposed, such as the Feature Pyramid Network (FPN) \cite{lin2017feature} and subsequent work \cite{Tan2019, ghiasi2019fpn, Wangc}, which aggregates
features from the backbone at different levels in order to create semantically rich 
feature maps at each level. %
Recently, FCOS \cite{Tian2019}, an anchor-free detection neural network architecture, has been 
developed. By leveraging the pyramid structure given by the FPN, 
FCOS removes the problem of anchor box assignment, and thus streamlines the object detection
task with a one-stage, proposal-free, anchor-free framework.

\paragraph{Instance segmentation neural networks}

Instance segmentation neural networks were pioneered by Mask-RCNN \cite{he2017mask},
a two-stage approach based on \cite{Girshick2014}. 
A significant number of architectures have been developed \cite{Wang2020a, Wanga, chen2019hybrid, Liu_2018_CVPR, cao2019gcnet},
each bringing incremental improvements in popular benchmarks.
Some methods save storage space by regressing masks as low-dimensional embeddings, 
which are decoded using either a learned decoder \cite{Jetley_2017_CVPR, Zhang_2020_CVPR} 
which complexifies the model and its training,
or fixed functions such as the Discrete Cosine Transform \cite{Shen_2021_CVPR}, in which case
the encoding dimension is still quite large.
Nonetheless, mask-based instance segmentation methods all encounter
the drawbacks of the pixel-based representation, and are generally much
slower than their bounding box-based counterparts.

\paragraph{Contour regression methods}

Recently, the community has explored new ways of regressing the boundaries of 
objects. These methods aim to alleviate the aforementioned issues of the 
pixel-based mask representation, while still offering better IoU with detected objects than
bounding box based architectures, which most of them take as a starting point.

Snake-based methods \cite{peng2020deep, Liu_2021_WACV} represent shapes as polygons in pixel coordinates and
deform them iteratively using circular convolutions. While these methods can reach good accuracy levels at reasonable speeds, 
they generally make use of a high number of contour points, which can be costly in terms of storage on embedded devices, 
especially compared to shape encoding methods described below.

ESE-Seg \cite{Xu2019} is based on the YOLOv3 detection neural network \cite{Redmon2018} and 
regresses an explicit shape encoding for each detected object in the form of either 
Chebyshev or Fourier coefficients and a center.
The contour of the object is modeled as a function $\rho(\theta), \theta \in [0,2\pi]$,
revolving around this center, as shown on the inset Figure. 
Corresponding ground truth coefficients 
\begin{wrapfigure}{r}{0.4\columnwidth}
\begin{center}
\includegraphics[trim={17mm 15mm 0mm 10mm },width=0.39\columnwidth]{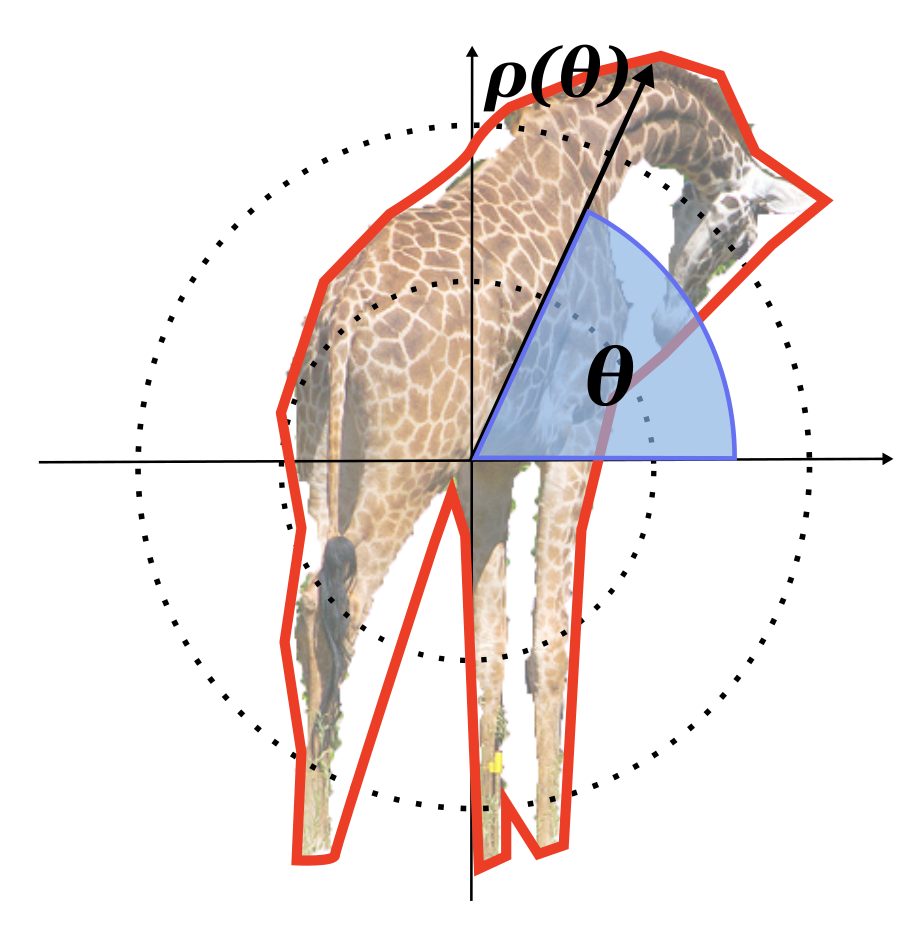}
\end{center}
\end{wrapfigure}
 have to be generated for each object of the training dataset. When used with Chebyshev coefficients, this representation is however not well suited to periodic functions, as we will demonstrate later. Moreover, this method is only able to fully represent star-shaped domains and thus cannot be used to regress less regular shapes.

In PolarMask \cite{Xie2019}, the authors modify the FCOS \cite{Tian2019} architecture
in order to regress the polar coordinates of the contour points directly, \ie the values of the function $\rho$. Again, specific ground truth labels have to be created
for each object, with a fixed number of rays cast from its center, 
and only star-shaped objects can be regressed, since only one coordinate
is regressed for each ray.

FourierNet \cite{Benbarka2020}, also based on FCOS, simplifies the contour regression
problem by using the Inverse Fast Fourier Transform (IFFT) as a differentiable shape
decoder. 
Thus, the neural network regresses Fourier coefficients, but the output is
directly compared to ground truth polygons. The authors use either polar or Cartesian
coordinates. 
In the first case, one set of complex coefficients is used to 
\begin{wrapfigure}{r}{0.25\columnwidth}
\begin{center}
\includegraphics[trim={15mm 5mm 0mm 20mm },width=0.25\columnwidth]{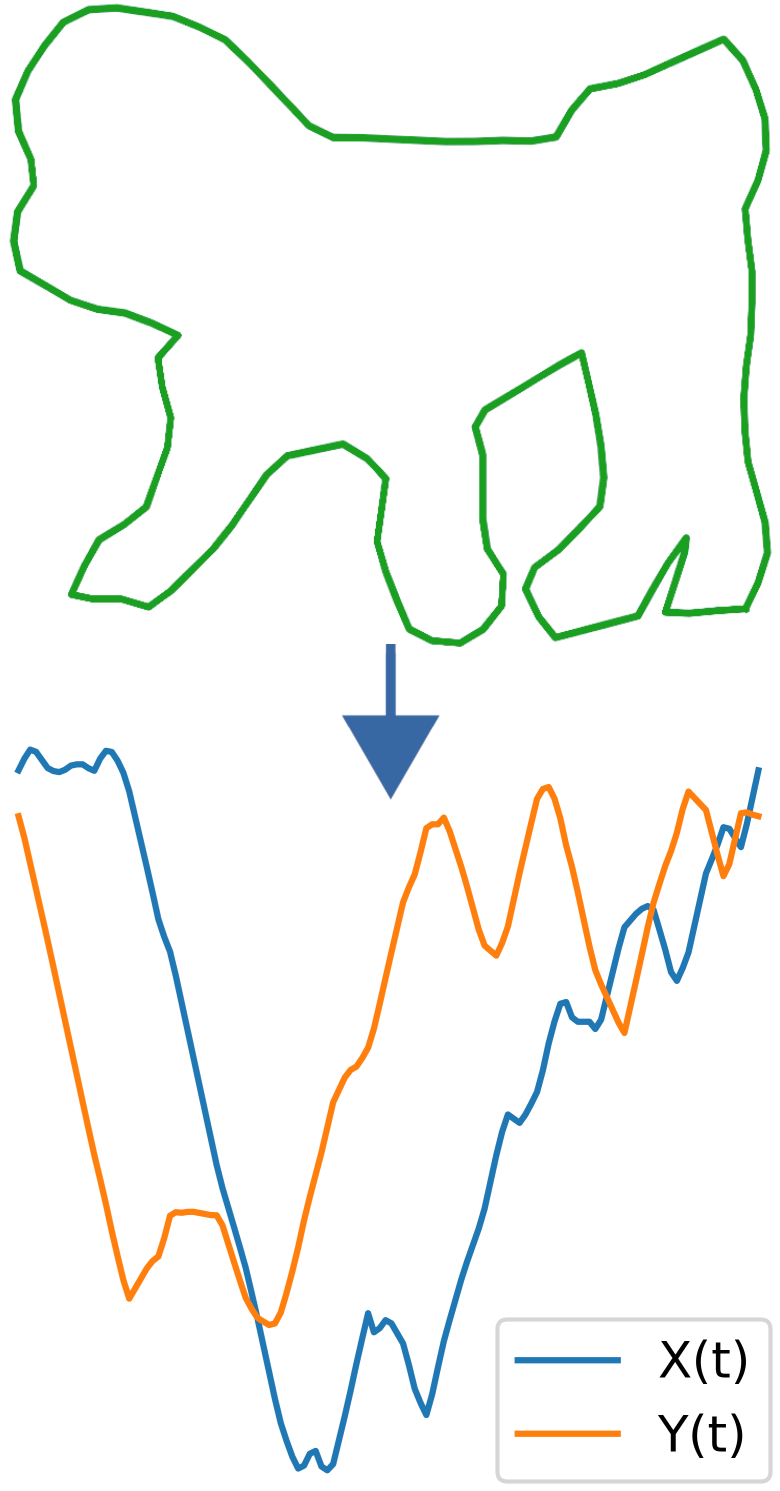}
\end{center}
\end{wrapfigure}
find $\rho$, 
and the regressed shapes have to be star-shaped. 
In the Cartesian case, the contour
is seen as two functions, $X(t)$ and $Y(t)$ as shown on the inset Figure. %
Two sets of
complex coefficients are used (one for each) to model these functions.
While the authors manage to get good looking results in the polar case, their Cartesian
version does not perform as well quantitatively, and
fails to produce coherent shapes when using a high number of Fourier coefficients.

\section{Method}

Our network architecture is based on the FCOS detection neural network \cite{Tian2019}. %
Like most recent detection architectures, FCOS is composed of a 
backbone, a neck (feature pyramid) and several detection heads with shared weights. %
As backbones, we use ResNet-50 \cite{He2015}, ResNeXt-101 \cite{Xie2017} and the lightweight DarkNet-53 \cite{Redmon2018},
in order to address different performance targets.
We run experiments with both the classical FPN neck from
\cite{lin2017feature}, as well as the more recent 
FPN-CARAFE \cite{Wangc}, which provides a small accuracy improvement.

\paragraph{Choice of shape representation}

In \cite{Benbarka2020, Xu2019, Xie2019}, the shape of an object is represented as a function
$\rho(\theta)$ with $\theta \in [0, 2\pi]$, and thus revolve around a center, which is either regressed
along with the coefficients in \cite{Xu2019}, or is the center of output "cells" from FCOS in \cite{Benbarka2020, Xie2019}. 
$\rho$ is a $2\pi$-periodic function measuring the distance of each point of a contour to this center. By contrast, we choose to model the shape of objects with a more generic 
complex periodic function of the form $C(t) = X(t) + iY(t)$ with $t \in [0,1]$.

\paragraph{On Chebyshev coefficients}

The use of Chebyshev coefficients \cite{Xu2019} is not well 
suited to the regression of continuous periodic functions through a neural network - in this case, the $\rho$ function. 
Indeed, we demonstrate in the appendix that $\rho$ is continuous periodic, \ie $\rho(0) = \rho(2\pi)$, if and only if the regressed Chebyshev coefficients $\alpha_i$ satisfy $\sum_{k=0}^{N/2} \alpha_{2k+1} = 0$.
It is unlikely that a neural network can learn this exact relation, even if optimized as a loss during training. 
Hence, discontinuities can be observed in nearly every shape regressed in this manner, as seen in Figure~11 of \cite{Xu2019} reproduced in appendix.
A similar relationship can be found using the derivatives of Chebyshev polynomials
if we also want shapes to be smooth, which should be the case for most objects.

\begin{figure*}[h!]
    \centering
    \includegraphics[trim={0mm 16mm 0mm 5mm},width=0.95\textwidth]{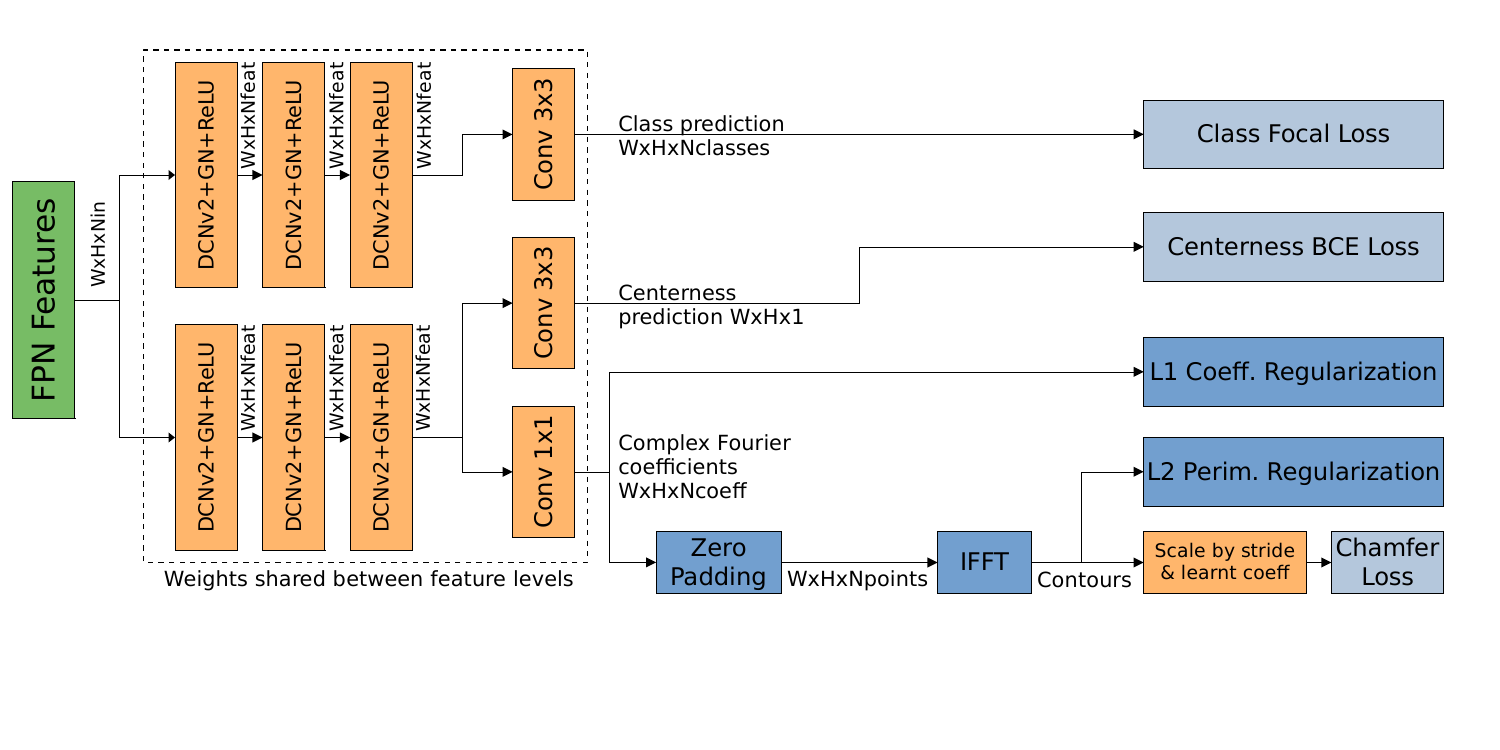}
    \caption{Our SCR head, which is applied to each output feature level of the FPN. Learned operators are in orange. Fixed operators are in blue.
    One branch of the head is used for the class prediction. The other branch is used for both centerness prediction and Fourier coefficient
    regression. All weights are shared between feature levels except the final scaling coefficient. 
    An $L^1$ regularization is applied to the Fourier coefficients to direct training towards the regression of simpler shapes. 
    An $L^2$ regularization is used on the perimeter of the regressed contours to encourage the regression of shapes that circle the object only once. }
    \label{fig:head}
\end{figure*}

\paragraph{Contours as a complex function, and Fourier series}

This drawback of Chebyshev 
polynomials steers us toward the Fourier decomposition, 
which is better suited to our purposes simply due to its conception,
since any set of Fourier coefficients will always represent a continuous smooth periodic function.

Our single complex function representation has several advantages over using two real functions as in \cite{Benbarka2020}: 
not only does it halve the number of IFFT calls in the shape decoding stage; it also ties these 
coefficients together as they represent a single function, which might help
during the training of the neural network, as backpropagation occurs through
a single IFFT call.

In order to evaluate how many complex coefficients are needed to represent 
shapes 
accurately, we interpolate all polygons of the COCO 2017 validation set %
using the %
scheme detailed in section \ref{interp}, we then compute the FFT
and its inverse after zeroing a certain number of coefficients,
and finally compare the result to the original polygons using the Chamfer Distance
\begin{wrapfigure}{l}{0.45\columnwidth}
\begin{center}
\includegraphics[trim={0mm 15mm 10mm 15mm },width=0.45\columnwidth]{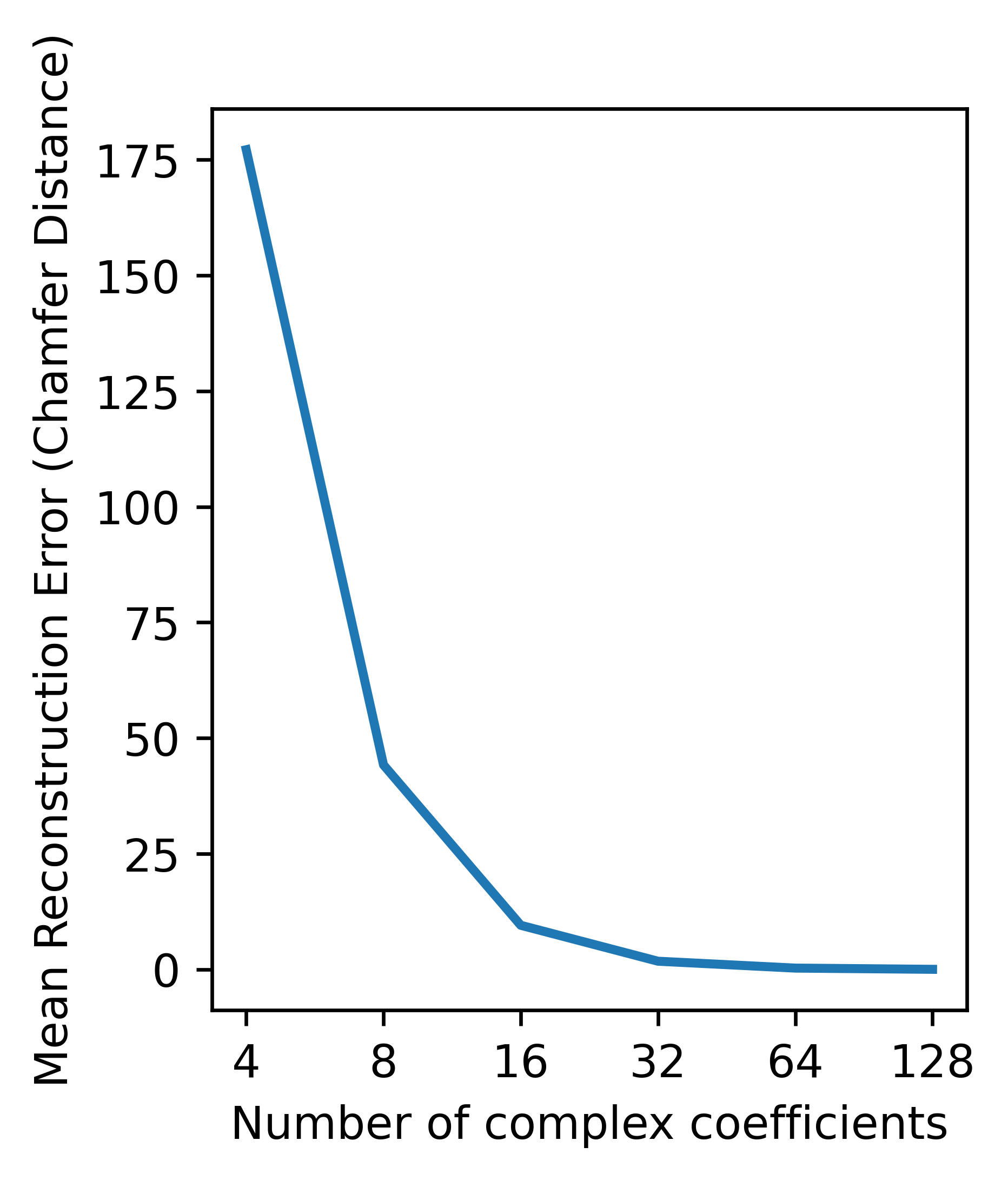}
\end{center}
\end{wrapfigure}
as reconstruction error. %
The inset figure shows 
the quickly diminishing returns of increasing the number of coefficients, 
highlighting the fact that only
a small number of them suffice to represent a shape accurately. We decide
to run our experiments using { 8} to 32 complex coefficients.

\paragraph{SCR head}

The  SCR head, shown in Figure \ref{fig:head}, is based on the FourierNet \cite{Benbarka2020} head.
For both the class and contour regression branches, we make use of 3 deformable convolutions \cite{Zhu} with shared
weights among feature levels. Thanks to the use of the complex contour representation, we only require
one call to the IFFT for shape decoding. We add two regularization losses
which we will detail further below. The first one is applied directly to the regressed coefficients. 
The second one is applied to the decoded point coordinates before they are scaled according to the 
stride of the feature level and a learnable scaling coefficient.

\paragraph{Loss function}

For training  SCR, our loss function is defined as a sum of different terms:
\begin{equation}
    \mathcal{L} = \loss{CD} + \loss{cent} + \loss{cls} + \loss{perim} + \loss{coeff}
\end{equation}
where $\loss{CD}$ is a polygon regression loss based on a symmetrized Chamfer Distance (similarly to \cite{Benbarka2020}), $\loss{cent}$ is the centerness loss
from FCOS \cite{Tian2019} (binary cross-entropy), and $\loss{cls}$ is the Focal Loss for classification
from RetinaNet \cite{Lin2017}. $\loss{perim}$ and $\loss{coeff}$  are regularization terms on the shape perimeter and on Fourier coefficient respectively. Both are detailed below.

In contrast to \cite{Benbarka2020}, note that we do not use any bounding box loss function as bounding boxes can be inferred from the regressed polygons and do not need to be learned separately.

\paragraph{L2 Perimeter regularization}

\label{perimeter_penalty}
The use of the Chamfer distance loss as the main polygon regression loss has 
a major drawback that needs to be addressed: since ground truth points 
are only compared to the closest regressed point, the network does not necessarily
learn to regress contours that go around the object only once. Indeed, circling the object
multiple times is not penalized in $\loss{CD}$.
One way to alleviate this problem, as done in \cite{Benbarka2020}, is to
first perform a warm-up of the shape regression with an $L^1$ loss, so that each
output point is already tied to specific ground truth points. However,
after being trained with Chamfer loss, their network still tries to regress
shapes in an overly complex way, which creates lots of self-intersections in the output polygon, as seen
in our comparison in section \ref{expl1} (Figure \ref{fig:messi}).

We instead choose to apply a $L^2$ perimeter-based regularization to every output contour,
even the ones that have not been assigned to any object. This, while also simplifying the training process, encourages simpler
shapes that go around the object once and have a minimal number of self-intersections:

\begin{equation}
    \displaystyle \loss{perim} = \lambdaloss{perim} \sqrt{ \sum_{i=0}^{N} (\mathbf{x}_i - \mathbf{x}_{i+1})^2}
\end{equation}

The $\lambdaloss{perim}$ coefficient can be set to a relatively low number %
in order to avoid influencing training too much. 
Setting it %
too high will result in small round-ish shapes.
It can be decayed over time or set to zero after $\loss{perim}$ reaches
a steady state, in order to retrieve more complex shapes while retaining the effect of regularization, 
as shown in our experimental results (Figure \ref{fig:decay_result}).

\paragraph{L1 Fourier coefficient regularization}

While representing contours using Fourier coefficients is already more 
storage-efficient than using masks, we may want to further reduce the 
storage space needs by ignoring very small coefficients (\ie setting them
to zero after regression). Thus, it would be interesting to enforce 
sparsity in the coefficients, in order to have as many of them be close to zero
as possible.

In order to favor sparsity of the Fourier coefficients $F_i$, we use an $L^1$
regularization term:

\begin{equation}
\displaystyle \loss{coeff} = \frac{\lambdaloss{coeff}}{N_c} \sum_{i=-n, i\notin \{-1,0,1\}}^{n} |F_i|
\end{equation}

Where $\lambdaloss{coeff}$ is an adjustable parameter (500.0 in our experiments).
It has to be noted that we do not apply this penalty to the -1, 0 and 1 frequency 
coefficients, since these coefficients will have to be quite large in most cases.
Having lots of Fourier coefficients close to zero, especially high frequency ones,
also helps the network create simpler and smoother contours, 
as shown in section \ref{expl1} (Figure \ref{fig:ablation_penalties}).

\paragraph{Polygon ground truth processing}
\label{interp}

Existing works \cite{Benbarka2020,Xie2019,Xu2019} typically convert the ground truth polygons to polar coordinates with respect to a center. This is done by casting a number of rays at regular angular intervals from this center 
to the farthest contour point. This process can hardly be done on-the-fly
without slowing the data loading pipeline.
We also note that
the distance between two adjacent points created in this manner
can vary greatly (\eg in oblong objects).
We believe that having a high number of evenly spaced out ground truth points is 
better for the stability of the Chamfer Distance loss. 
Thus, our interpolation is done in a "constant spacing" fashion for 
each ground truth polygon. %
Since the original ground truth polygons vary in number of points,
\begin{wrapfigure}{r}{0.45\columnwidth}
\begin{center}
\includegraphics[trim={13mm 15mm 3mm 12mm },width=0.45\columnwidth]{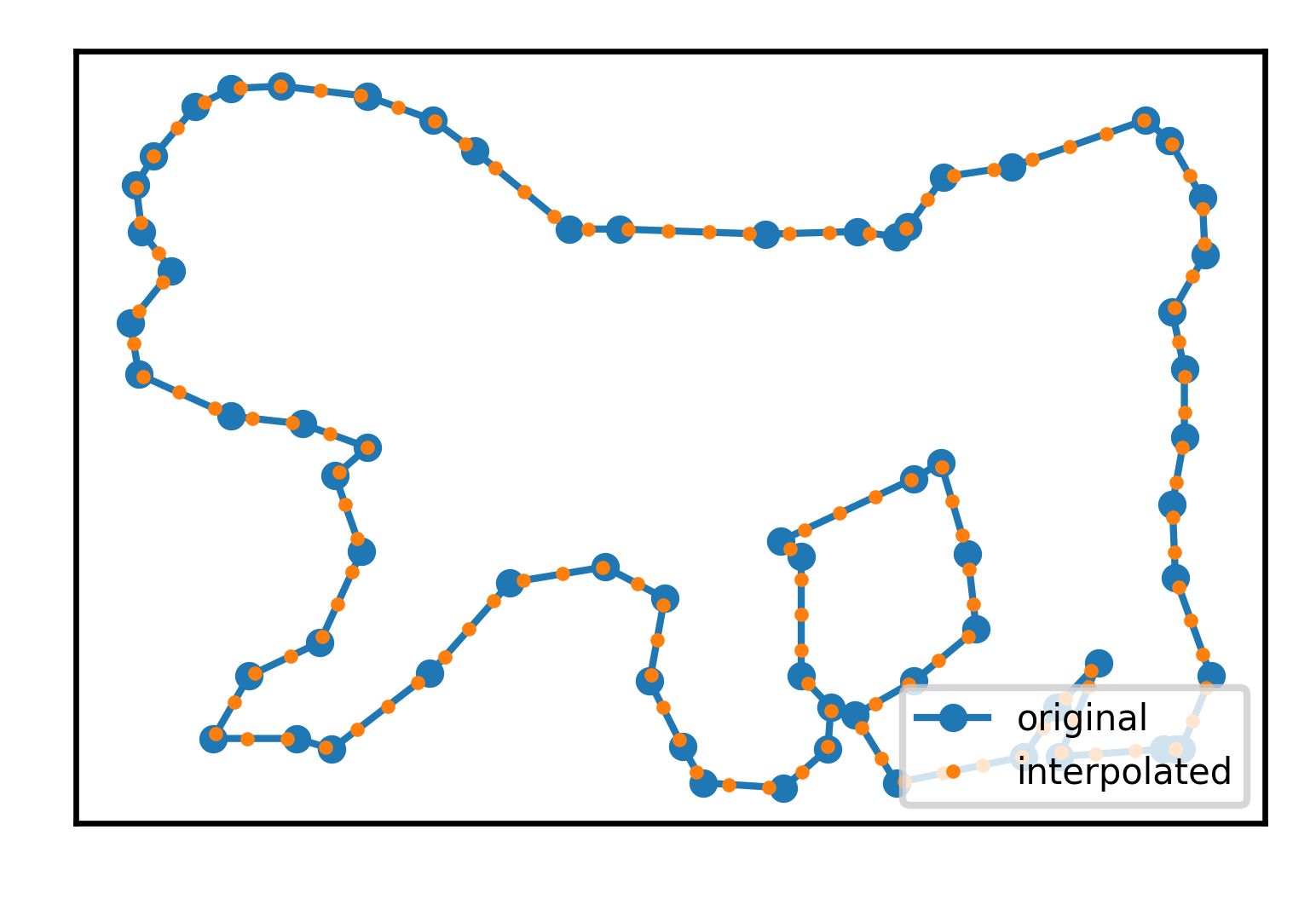}
\end{center}
\end{wrapfigure}
we interpolate them such that the number of ground truth points $N_p$
matches the number of regressed points. This is easily done
on the fly in the data loading pipeline and does not incur any
performance loss. %

\section{Experiments}
\label{expl1}

We implement our ResNet-50, ResNeXt-101 and DarkNet-53 based models using the MMDetection 2.7.0 object detection toolbox \cite{mmdetection}, based
on MMCV 1.2.1 for easy and fair comparison with other methods.
The underlying framework is PyTorch 1.7.0 with CUDA 10.1.
We make use of 
the updated FFT package from PyTorch 1.7.0 which allows the use of complex numbers.
We use the Chamfer Distance implementation from PyTorch3D 0.3.0 \cite{ravi2020pytorch3d}. 
The interpolation scheme described in \ref{interp} is implemented as a module in the MMDetection
data loading pipeline using the "interp" module from Numpy.
Notably, due to software incompatibilities during the conversion of our DarkNet-53 model for low-power hardware targets,
we had to simplify the architecture by removing deformable convolutions and CARAFE. %
In order to evaluate our method and easily compare it to previous
works, we choose to work on the widely used COCO 2017 Instance Segmentation dataset \cite{Lin2014}.

Our networks are trained for 38 epochs using SGD with momentum (0.9), with an initial learning
rate of 0.0003, batch-size of 16 (2 images per GPU), 
gradient clipping at a maximum $L^2$ norm of 45, and employ of a 500 step warm-up with a ratio of 0.001 applied to the learning rate.
We use the multiscale training feature found in MMDetection to train with image heights of 640 and 800 pixels.
We balanced our loss function terms through trial-and-error, and settled on the following values: 
$\lambdaloss{cls}$=1.0, $\lambdaloss{cent}$=1.0, $\lambdaloss{CD}$=1.0, $\lambdaloss{perim}$=0.01, $\lambdaloss{coeff}$=500.0.

\begin{table}[h!]
    \centering
    \setlength\tabcolsep{3pt}
    \begin{tabular}{c|c|c|c|c|c}
         Name-Backbone & DCNv2 & CARAFE & $N_{in}$ & $N_{feat}$ & Time  \\ \hline
          SCR-D53 & No & No & 128 & 128 & 20h \\
          SCR-R50 & Yes & Yes & 256 & 256 & 40h \\
          SCR-X101 & Yes & Yes & 256 & 256 & 80h \\
    \end{tabular}
    \caption{Characteristics of our main networks. Variations are further explored in our ablation study.
    DCN refers to the use of Deformable Convolution v2 \cite{Zhu} layers in the head. CARAFE refers to the use of the
    FPN-CARAFE \cite{Wangc} neck. Training time on 8 Tesla V100 GPUs.}
    \label{tab:hyperp}
\end{table}

\paragraph{Qualitative results}

Figure \ref{fig:teaser} shows a selection of results from the COCO 2017 test-dev dataset,
computed using our smallest DarkNet-53 based network.
Our method properly regresses smooth artifact-free contours in these scenes varying in
number of objects, classes and shapes. In particular, non-star-shaped objects can be accurately represented. An extended selection of results is
provided in appendix.

\paragraph{Quantitative comparison}

{ 
We compare our method to existing shape encoding methods \cite{Benbarka2020, Xu2019, Xie2019}, as well as state-of-the-art snake-based \cite{peng2020deep, Liu_2021_WACV} and mask-based methods \cite{he2017mask, Liu_2018_CVPR, Wang2020a}. 
The evaluation is done using three metrics: mAP for accuracy (as defined in \cite{Lin2014}), FPS (frames per second) reported on a single Nvidia GTX 1080Ti GPU for speed and 
memory usage per regressed object in bits, which we call \textit{SEC} (for Shape Encoding Complexity).
Run time and object size in memory or storage are important factors for embedded hardware applications, 
such as real-time on-board 
change detection. Thus, we propose OES (Overall Efficiency Score), %
which is simply the multiplication of these three scores and thus 
defined as $OES = 100 \times mAP \times FPS \times SEC^{-1}$. This metric represents the trade-off between having slow but accurate models outputting detailed shapes, and having faster models
outputting simpler shapes.

The comparisons against other shape encoding methods are shown in Table \ref{tab:coco101} for the ResNeXt-101 backbone and \ref{tab:coco53} for the
DarkNet-53 backbone.
We observe that with the same backbone and at a similar number of coefficients, our method achieves good accuracy compared to other shape encoding methods.

In Table \ref{tab:coco101}, our method has a competitive efficiency score (OES).
While \cite{Benbarka2020} is a little bit faster, our method reaches a better accuracy.
By contrast, PolarMask \cite{Xie2019} can achieve higher mAP at the cost of a higher number of coefficients, and thus has a lower overall efficiency.

Table \ref{tab:coco53} shows that our method is faster and more efficient than ESE-Seg \cite{Xu2019}.
In addition, even though its mAP score is a little bit lower, 
our method will not visually suffer from the drawbacks of Chebyshev coefficients detailed in section 3.

\begin{table}[h]
    \centering
    \setlength\tabcolsep{3pt}
    \begin{tabular}{c|c|c|c|c|c}
        Method &  $\Ntxt{coeff} $ & mAP & FPS &  SEC &  OES \\ \hline %
        PolarMask \cite{Xie2019}& 36 & \textbf{32.9} &  4.1 & 1152 & 11.7  \\ 
        FourierNet-Cartesian \cite{Benbarka2020} & 16  & 22.9 &  \textbf{4.9} & \textbf{512} & 21.9 \\
        FourierNet \cite{Benbarka2020} & 16 & 23.3 &  \textbf{4.9} & \textbf{512} & 22.3 \\ \hline
        \textbf{Ours} (SCR-RX101)  & 16  & 27.3 & 4.2 & \textbf{512} & \textbf{22.4} \\
        \end{tabular}
    \caption{COCO 2017 test-dev results against other shape encoding methods based on ResNeXt-101 with an image height of 800 pixels. $\Ntxt{coeff}$ in real numbers for shape encodings: one complex coefficient counts as two. 
    SEC: size
    of a single detected object in memory (bits). 
    OES: Overall Efficiency Score.%
    }
        \label{tab:coco101}
\end{table}

\begin{table}[h]
    \centering
    \begin{tabular}{c|c|c|c|c|c}
        Method &  $\Ntxt{coeff} $ & mAP & FPS &  SEC &  OES \\ \hline %
        ESE-Seg\cite{Xu2019} & 20 & \textbf{21.6} & 38.5 & 640 & 130 \\%44.4 \\
         \textbf{Ours} (SCR-D53) & \textbf{16}  & 21.2 & \textbf{39.1} & \textbf{512} & \textbf{162} \\%\textbf{44.8} \\%\textbf{40.6} \\
        \end{tabular}
    \caption{COCO 2017 val results for models based on DarkNet-53 with an image height of 416 pixels. $\Ntxt{coeff}$ in real numbers for shape encodings: one complex coefficient counts as two. 
    SEC (Shape Encoding Complexity): size
    of a single detected object in memory (bits). 
    OES: Overall Efficiency Score.%
    }
        \label{tab:coco53}
\end{table} 

The comparison against other types of approaches is shown in Table \ref{tab:coco50}.
Snake-based methods typically offer a better accuracy than shape encoding methods, however their overall efficiency suffers from a high shape encoding 
complexity and are slower than our method when using the same processing size and backbone.
Mask-based methods reach very high mAP scores, but are also hindered by their speed and SEC and thus generally have a lower overall efficiency.

\begin{table}[h]
    \centering
    \setlength\tabcolsep{3pt}
    \begin{tabular}{c|c|c|c|c|c|c}
        &Method &  $\Ntxt{coeff} $ & mAP & FPS &  SEC &  OES \\ \hline %
        \parbox[t]{2mm}{\multirow{2}{*}{\rotatebox[origin=c]{90}{\textit{Snake}}}} &  DeepSnake \cite{peng2020deep} & 256 & 31.0 & 6.68* & 4096 & 5 \\
        
        &  DANCE \cite{Liu_2021_WACV}  & 196 & 34.6 &  7.6 & 3136 & 8 \\ \hline
        
        \parbox[t]{2mm}{\multirow{3}{*}{\rotatebox[origin=c]{90}{\textit{Mask}}}} &  Mask R-CNN \cite{he2017mask} & 784 & 33.6 &  10.8 & 784 & 46 \\
        &  PANet \cite{Liu_2018_CVPR} & 784   & 38.2 &  4.5 & 784 & 22 \\ %
        &  SOLOv2 \cite{Wang2020a} & 64000  & \textbf{38.8} &  \textbf{12.4} & 64000 & 0.75\\ \hline %
        &\textbf{Ours} (SCR-R50) &   \textbf{16}  & 24.2 & 10.7 & \textbf{512} & \textbf{51} \\ 
        \end{tabular}
    \caption{COCO 2017 test-dev results against mask and snake-based methods with a ResNet-50 backbone and an image height of 800 pixels. $\Ntxt{coeff}$ in real numbers for shape encodings: one complex coefficient counts as two. 
    SEC: size
    of a single detected object in memory (bits). 
    For snake-based methods, we assume pixel coordinates are stored as 16-bit integers.
    For mask-based methods, we assume masks are binary.
    OES: Overall Efficiency Score.
     *: Number extrapolated from \cite{Liu_2021_WACV}
    }
        \label{tab:coco50}
\end{table}

While our method does not necessarily give the best score for each of the evaluation metrics, it offers a good compromise between them. In particular, the OES scores obtained by our method
 is competitive compared to existing methods under the same backbone.

}

{ 
}

\begin{figure}[h!]
    \centering
    
    \label{fig:l_perim_decay}
    \includegraphics[width=0.49\columnwidth]{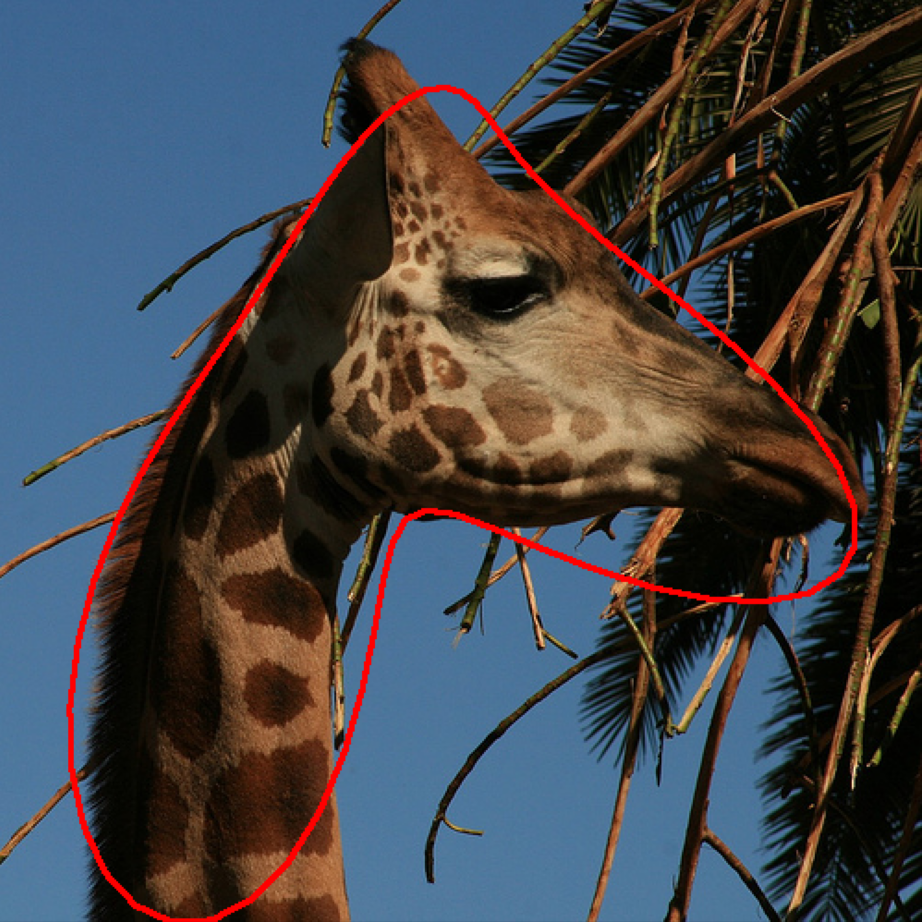}
    \includegraphics[width=0.49\columnwidth]{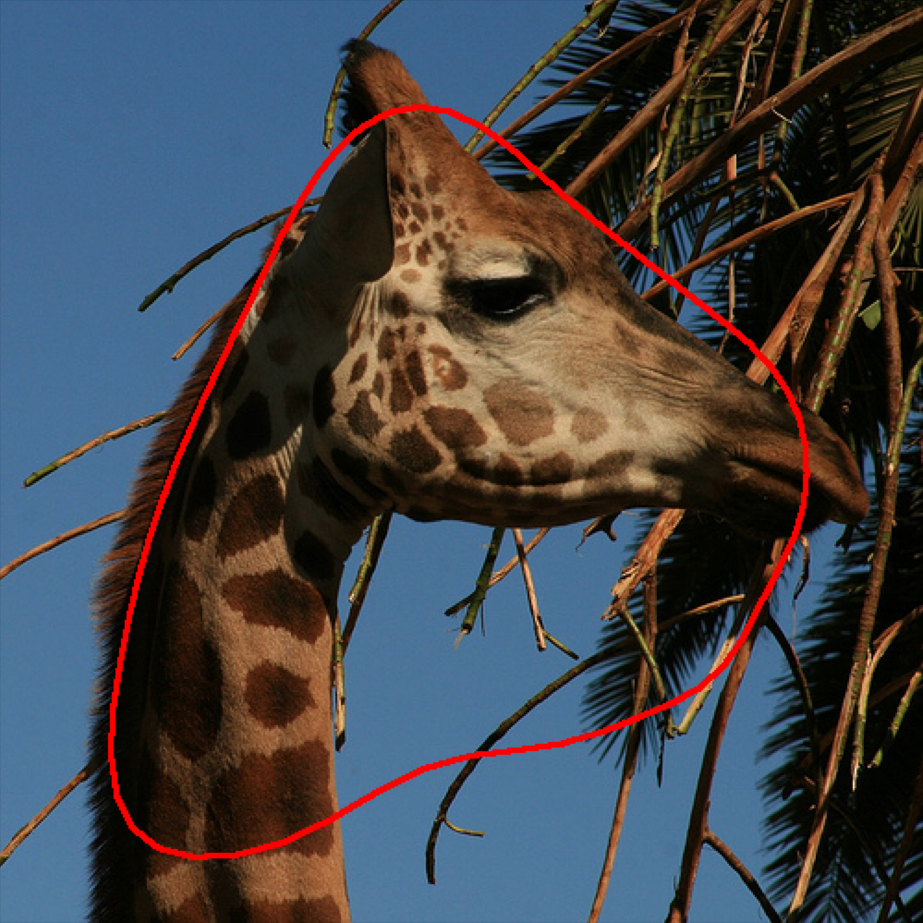}
    \caption{ SCR-D53 result with (left) and without perimeter penalty decay (right). 
    The perimeter regularization loss coefficient $\lambdaloss{perim}$ is set to zero once $\loss{perim}$ reaches steady state (2000 iterations here), thus leading to a more detailed shape while retaining the effect of regularization.}
    \label{fig:decay_result}
\end{figure}
\paragraph{Effect of perimeter penalty decay}

As discussed in section \ref{perimeter_penalty}, it is possible to set $\lambdaloss{perim}$ to zero after $\loss{perim}$ stabilizes. %
Removing the perimeter regularization after a few iterations
 (2000 here) leads to more complex and accurate contours, while retaining the 
effect of regularization. On figure \ref{fig:decay_result},
the network learns to properly follow both front and back legs of the giraffe (left), which is not the 
case without perimeter penalty decay (right). 
This is also a typical example of non-star-shaped contour
that cannot be regressed by previous works that opt for a single polar function.
Setting $\lambdaloss{perim}=0$ right away, \ie removing $\loss{perim}$ entirely, will let the
network circle objects multiple times, leading to poor results, { as shown in our ablation study and on Figure \ref{fig:ablation_penalties}(a)}.

\begin{figure}[h!]
\includegraphics[width=\columnwidth]{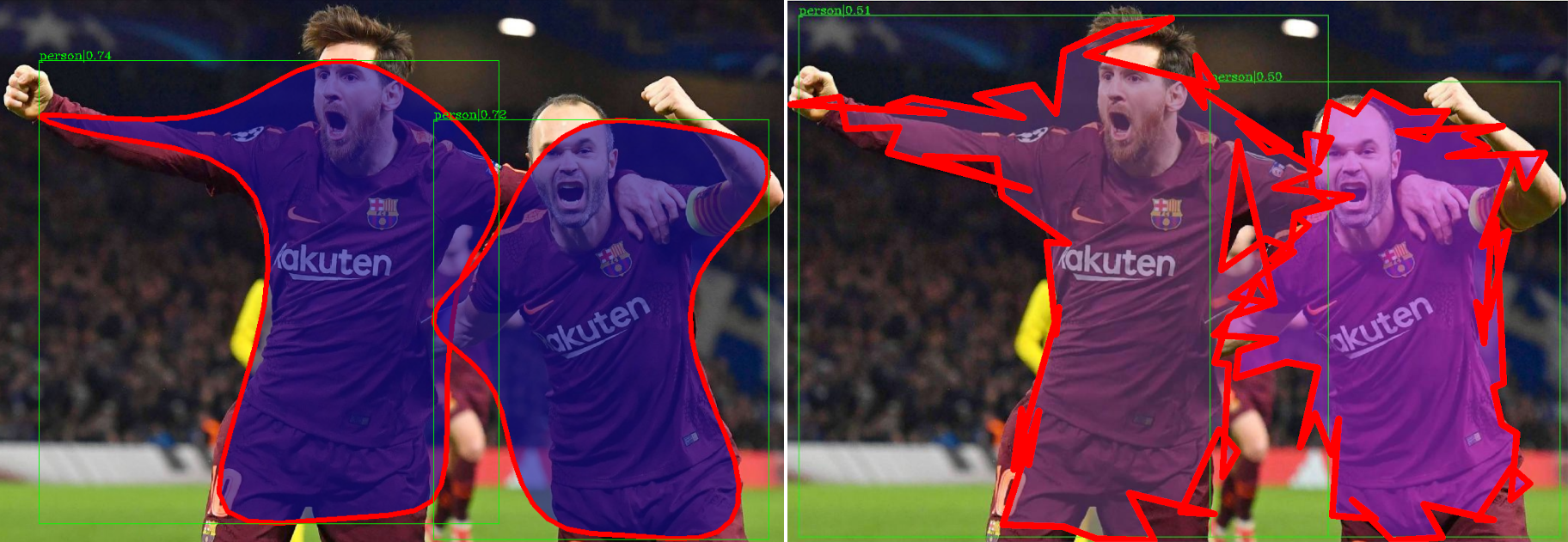}
\caption{Thanks to our coefficient regularization, our method \textbf{(left)} learns to regress smooth shapes compared to
the erratic ones regressed the cartesian version of \cite{Benbarka2020} \textbf{(right)}
when using a high number of coefficients (here 32).}
\label{fig:messi}
\end{figure}

\begin{figure*}[h!]
    \centering
     \setlength\tabcolsep{2pt}
    \begin{tabular}{ccc}
        \includegraphics[width=0.33\textwidth]{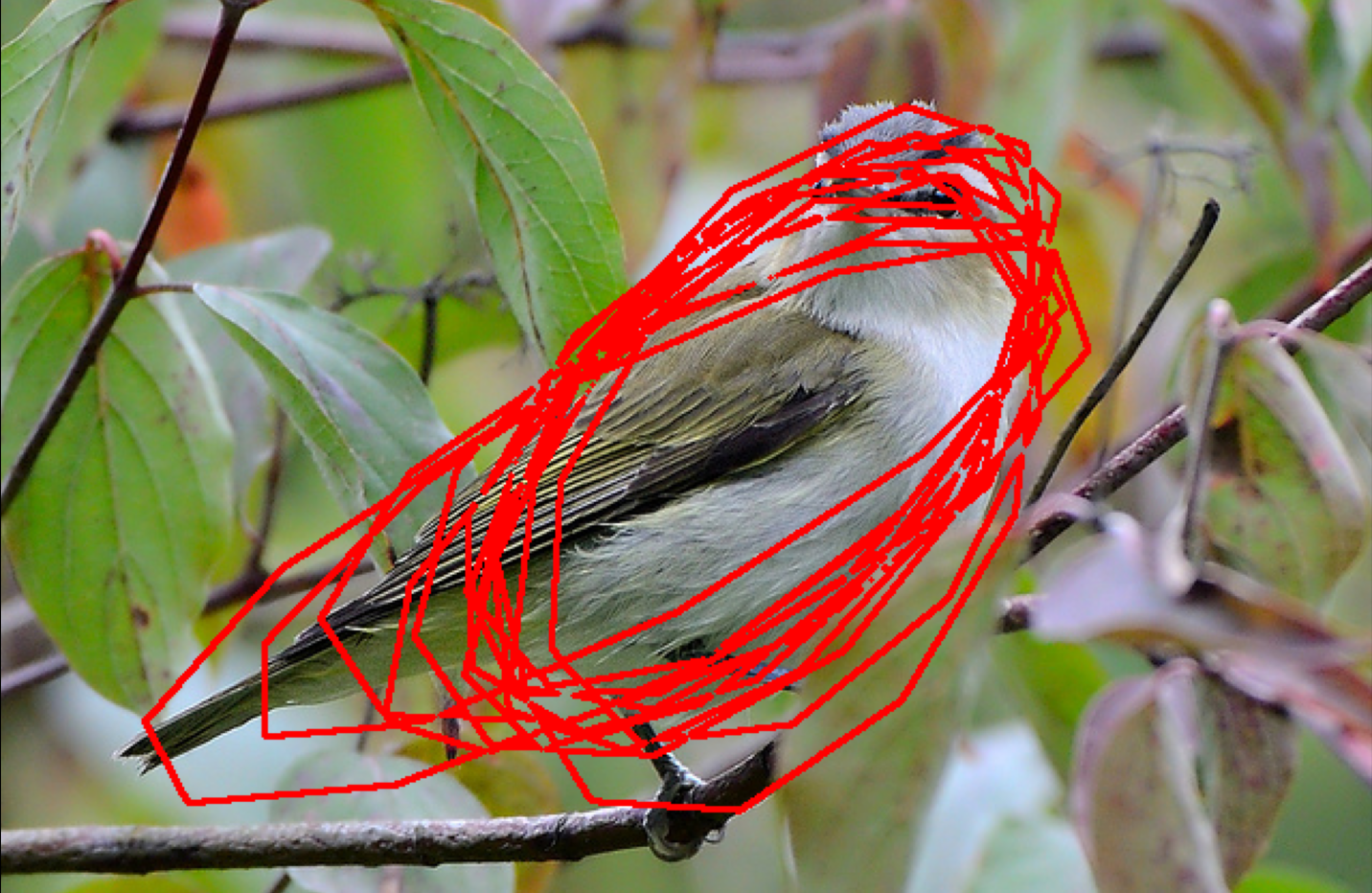}&
        \includegraphics[width=0.33\textwidth]{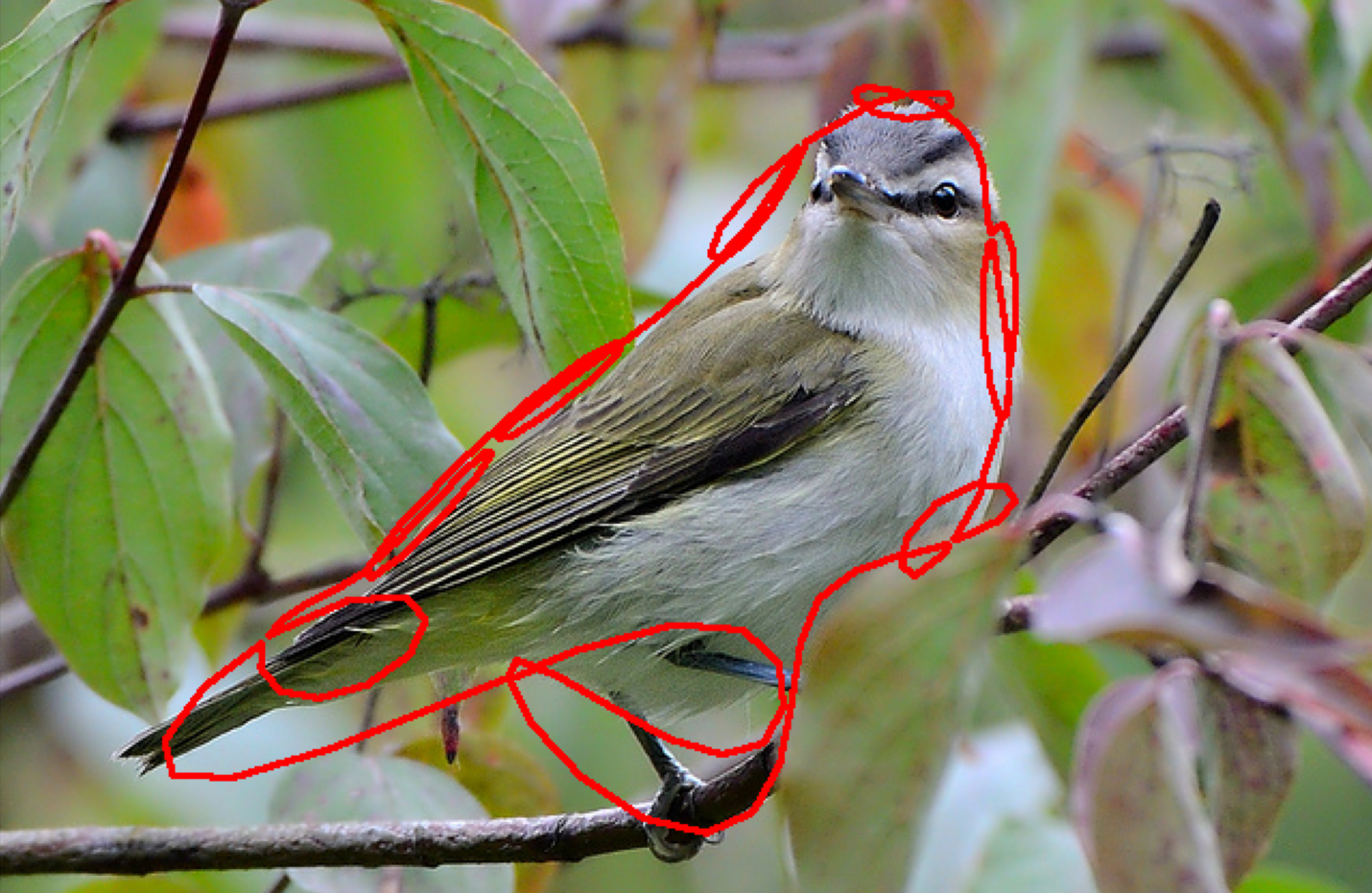}&
        \includegraphics[width=0.33\textwidth]{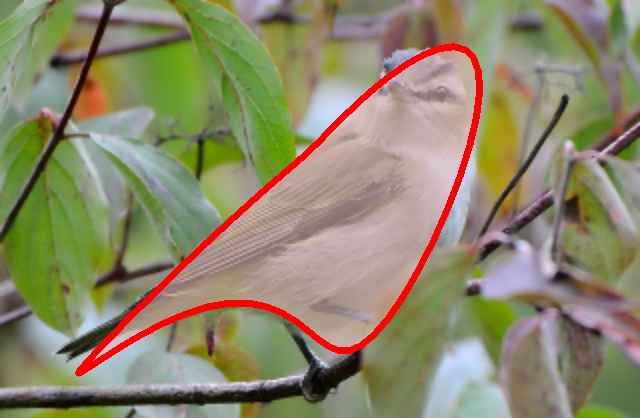} \\
        (a) Perim. reg. disabled & (b) Coeff. reg. disabled & (c) Both reg. enabled 
    \end{tabular}

    \caption{Disabling our perimeter regularization (a) lets the network circle the object multiple times, while 
    disabling the coefficient regularization (b) leads to erratic shapes with lots of self-intersections.
    When both regularizers are enabled (c), the network outputs smooth and accurate shapes with a low number of
    self-intersections, here zero.}
    \label{fig:ablation_penalties}
\end{figure*}

\newcommand{\cmark}{\ding{51}}%
\newcommand{\xmark}{\ding{55}}%

\begin{table*}[h!]
    \centering
     \setlength\tabcolsep{3pt}
    \begin{tabular}{c|c|c|c|c|c|c|c|c|c|c|c|c|c|c}
        Exp. & Backbone & Neck & Head & $\Ntxt{coeff} $ & $N_{pts}$ & H & Perim. reg & Coeff. reg. & mAP & AP$_{50}$ & AP$_{75}$ & AP$_{S}$ & AP$_{M}$ & AP$_{L}$ \\ \hline
        \#1 & Resnet-50 & FPN & DCN & 20 & 60 & 800 & \cmark & \cmark  & 22.4 & 43.0 & 21.0 & 10.8 & 24.2 & 29.7 \\  \hline 
        \#2 & Resnet-50 & CARAFE & DCN & 20 & 128 & 800 & \cmark  & \cmark  & 21.4 & 44.2 & 19.0 & 10.9 & 22.9 & 28.4 \\ \hline 
        \#3 &  &   &   & 20 &  &  & & & 23.8 & 45.0 & 22.4 & 11.6 & 25.6 & 31.9 \\  
        \#4 & Resnet-50 & CARAFE & DCN & 40 & 60 & 800 & \cmark  & \cmark  & 23.8 & 46.4 & 21.8 & 11.7 & 25.4 & 32.3 \\
        \#5 &  &   &   & 64 &  &   & & & 25.2 & 48.1 & 23.6 & 13.1 & 26.9 & 33.6 \\ \hline 
        \#6 & Resnet-50 & CARAFE & DCN & 20 & 60 & 360 & \cmark  & \cmark & 19.1 & 36.2 & 18.1 & 4.0 & 19.9 & 31.3 \\  
        \#7 & Resnet-50 & CARAFE & DCN & 64 & 60 & 360 & \cmark  & \cmark  & 18.9 & 37.3 & 17.0 & 4.3 & 19.5 & 30.9 \\ \hline 

        \#8 & &  &  &  &  & 320 & & & 17.6 & 33.8 & 16.4 & 1.5 & 17.0 & 32.1 \\
        \#9 & DarkNet-53 & FPN & - & 16 & 60 & 416 & \cmark  & \cmark  & 21.2 & 39.3 & 20.4 & 3.7 & 22.6 & 35.5 \\
        \#10 & &  &  &  &  & 608 &   &  & 23.9 & 42.2 & 24.0 & 6.2 & 27.5 & 36.3 \\ \hline
         \#11 & & & & & & & \cmark  &\xmark & 10.8 & 22.6 & 9.3 & 6.3 & 12.5 & 14.2 \\
         \#12 & Resnet-50 & CARAFE & DCN & 20 & 128 & 800 & \xmark & \cmark  & 5.8 & 14.5 & 3.5 & 3.1 & 6.6 & 7.8 \\
         \#13 & & & & & & &\xmark &\xmark  & 5.4 & 13.9 & 3.1 & 2.9 & 6.1 & 7.3  \\ \hline 
        \#14 & ResNeXt-101 & CARAFE & DCN & 20 & 60 & 800 & \cmark  & \cmark  & \textbf{27.2} & 50.3 & \textbf{26.4} & \textbf{14.0} & 29.4 & 35.4 \\ 
         \#15 & ResNeXt-101 & CARAFE & DCN & 64 & 60 & 800 & \cmark  & \cmark  & \textbf{27.2} & \textbf{50.9} & 26.1 & 13.9 & \textbf{29.5} & \textbf{35.9} \\ 
    \end{tabular}
    \caption{Impact of different design choices on our models on COCO test-dev. H = image height. 
    $\Ntxt{coeff}$ in real numbers (one complex coefficient counts as two). $N_{pts}$ = number of contour points in ground truth and after IFFT. 
    DCN = Deformable Convolution v2.}
    \label{tab:ablation}
\end{table*}

\paragraph{Fourier coefficient regularization}
The benefits of our coefficient regularization
can be 
observed on figure \ref{fig:messi},
where our network (left) learned to regress smooth contours compared to the %
erratic shape regressed by the Cartesian version of FourierNet \cite{Benbarka2020} (right), even when using a 
high number of coefficients. Figure \ref{fig:ablation_penalties}(b) shows that disabling 
our regularization leads to the same kind of behavior shown in \cite{Benbarka2020}.

\vspace{-1.5em}

\paragraph{Ablation study}
We now evaluate the impact of our design choices through an ablation study. The results
 are summarized in Table \ref{tab:ablation}. Numbers in parentheses { in the following paragraph} refer to corresponding lines of that Table.

\textbf{(\#2,\#3)} We trained the ResNet-50 based model with 60 and 128 contour points,
and found that for this set of hyper-parameters, there was a performance
degradation when using 128 points. We thus kept this parameter at 60,
a finding consistent with the literature.
\textbf{(\#1,\#3)} We show that the new FPN-CARAFE from \cite{Wangc} improves the 
scores of  SCR compared to the vanilla FPN, for which it is a
drop-in replacement. 
\textbf{(\#3,\#4,\#5; \#6,\#7; \#14,\#15)} Increasing the number of complex coefficients from 10 (20 real values) to 20 
made no significant difference, while increasing it further to 32 yielded
a small performance increase of 1.4 mAP. Depending on the requirements of the application,
this increase in detail might be worth the extra storage space. 
Note that it is still possible to regress a high number of coefficients, then
decide to set some of them to zero after the fact.
\textbf{(\#6,\#7,\#8,\#9)} We find that the models used on smaller image resolutions do not perform
very well on small objects (AP$_{S}$).
\textbf{(\#14,\#15)} Using a larger ResNeXt-101 backbone yields a small accuracy improvement, but not 
to the point where we would think it is worth the decrease in FPS and the up to two times longer training
times. \textbf{(\#10)} DarkNet-53, on the other hand, seems to perform as well as ResNet-50 at high resolutions, especially on
larger objects.
{ \textbf{(\#2,\#11,\#12,\#13)} Disabling the $L_1$ coefficient penalty leads to a very significant drop in mAP. This is probably due
to the erratic shapes regressed without it, as shown on Figure \ref{fig:ablation_penalties}. Disabling the $L_2$ perimeter penalty also significantly affects the scores, as the contours 
circle the objects multiple times and thus miss a lot of detail.}

\paragraph{Limitations}

\begin{figure}[h!]
\centering
\includegraphics[height=0.49\columnwidth]{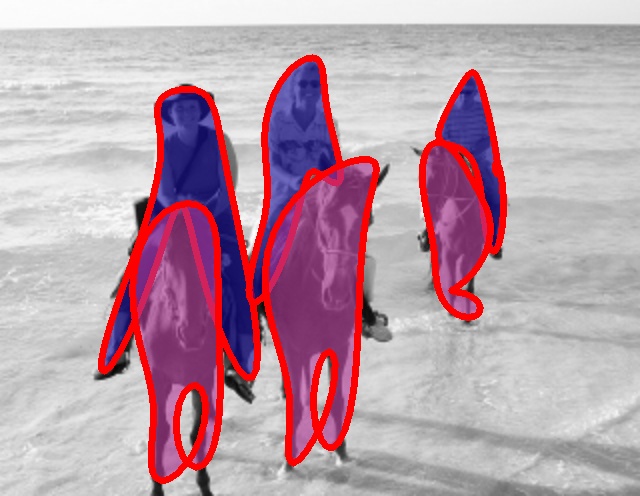}
\includegraphics[height=0.49\columnwidth]{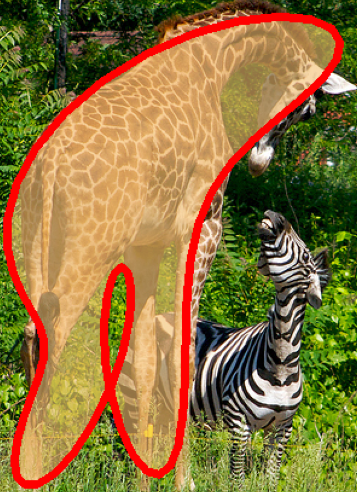}
\caption{Examples of self-intersections which can appear occasionally as they are not actively prevented.
The contours still follow the shape of the animals and are thus still somewhat usable.
}
\label{fig:limit}
\end{figure}

Our regularization terms tend to direct the training of our network 
towards 
simple shapes without actively preventing self-intersections. Thus, such errors
might appear occasionally.
On figure \ref{fig:limit}, for instance, we see that while the two leftmost horse contours seem
properly regressed, they each present one self-intersection. 
 Nevertheless, in that particular case, the masks given by these contours are still usable.
One other limitation is that by design, our method is more tailored towards free-form objects
than the regression of piece-wise linear shapes, for which polygon-based methods are better suited.

\paragraph{Performance benchmarks on low-power hardware}

We evaluate the speed of our fastest DarkNet-53 based
network on a variety of devices, ranging from powerful discrete
GPUs to low-power edge devices (Table \ref{tab:benchmark}).

Our first test platform is the Nvidia 
Jetson Nano, which is a popular,
affordable and widely available yet reasonably powerful 
family of development kits. 
Our method runs 
at 9.5 FPS on these boards, which is close enough to real-time to
be usable in many applications.
We also run the same benchmarks 
on the more powerful Jetson Xavier NX, %
featuring INT8
compute capabilities, thus yielding a significant performance boost by bringing
the throughput up to 115 FPS. It achieves 5.75 FPS/Watt, making it by far the 
most efficient combination.
The Intel Movidius Myriad X VPU is an ultra low-power AI
accelerator with a power consumption of only 1W. 
The Myriad VPUs have been used in ESA's $\phi$-sat-1 CubeSat,
as well as in smart security cameras,
UAVs and industrial machine vision equipment. We benchmark  SCR
on the Myriad X { using the imposed OpenVino toolkit} %
and achieve 2.8 FPS, which is very satisfactory
for such a low-power device.
As a point of comparison with other works, we also run this network on
an Nvidia GeForce GTX 1080Ti GPU, %
which is still one of the most popular discrete GPUs. 
For the sake of completeness, we also provide numbers on an Intel Xeon Silver 4114 CPU.

\begin{table}[h!]
\small
\setlength\tabcolsep{2pt}
    \centering
    \begin{tabular}{c|c|c|c|c|c|c|c|c}
        &Device & Power & Net. & H & Tool & Prec. & FPS & fps/W \\ \hline

         \parbox[t]{2mm}{\multirow{8}{*}{\rotatebox[origin=c]{90}{\textit{Low-power}}}} & Nano & 10W &  Ours & 320 & TensorRT & FP32 & 5.4 & 0.54 \\ 
        &&  & & & TensorRT & FP16 & 9.5 & 0.95 \\ \cline{2-9}
          
        &Nano & 10W &  Ours & 608 & TensorRT & FP32 & 1.72 & 0.17\\ 
        && & & & TensorRT & FP16 & 3.04 & 0.30 \\ \cline{2-9}
         
        &&  &  & & TensorRT & FP32 & 17.2 & 0.57 \\ 
        &Xavier NX & 30W &  Ours & 320 & TensorRT & FP16 & 66.7 & 2.22 \\
        && & & & TensorRT & INT8 & \textbf{115} & \textbf{5.75} \\ \cline{2-9} 
         
        &Myriad X & 1.5W &  Ours & 320 & OpenVino & FP16 & 2.8 & 1.87 \\ \hline  \hline  
        \parbox[t]{2mm}{\multirow{7}{*}{\rotatebox[origin=c]{90}{\textit{High-power}}}} &Xeon 4114 & 85W &  Ours & 608 & OpenVino & FP16 & 5.3 & 0.06 \\  \cline{2-9}
        
        &1080Ti & 250W  &  Ours & 320 & MMDet & FP32 & 40.6 & 0.16 \\
        &1080Ti & 250W  &  Ours & 320 & TensorRT & FP32 & \textbf{186.6} & 0.74  \\ \cline{2-9} 
        &1080Ti & 250W  &  Ours & 608 & MMDet & FP32 & 32.8 & 0.13 \\
        &1080Ti & 250W  &  Ours & 608 & TensorRT & FP32 & 68.2 & 0.27 \\ \cline{2-9} \cline{2-9} 
        &1080Ti & 250W  &  Ours & 416 & MMDet & FP32 & 39.1 & 0.15 \\ \cline{2-9} 
        &1080Ti & 250W & \cite{Xu2019} & 416 & PyTorch & FP32 & 38.5 & 0.15 \\ %
        & 1080Ti & 250W & \cite{Benbarka2020} & 800 & MMDet & FP32 &  4.9 &  0.02 \\
        & 1080Ti & 250W & \cite{Benbarka2020} & 360 & MMDet & FP32 &  16.5 &  0.07 \\ 
    \end{tabular}
    \caption{Throughput of  SCR with Darknet-53 backbone on a wide range of devices with varying levels of power.
    H = image height.}
    \label{tab:benchmark}
\end{table}

\section{Conclusion}

In this paper, we proposed  SCR, a method that captures object contours with a compact resolution-free representation based on a low number of Fourier coefficients.
Using a complex representation and geometric priors, 
we were able to observe qualitative and quantitative improvements in the regressed shapes over previous methods.

We also implemented a downsized version of our model which, while retaining
competitive accuracy, is able to run on a wide range of low-power hardware at 
very reasonable speeds (up to 115 FPS on a 30W device), 
and is therefore suitable for edge computing applications such as on-board processing of UAV or satellite images, autonomous vehicles and robotics.

Future work will include testing and evaluating  SCR on other applicative scenarios, such as autonomous driving and remote sensing. 
Matching the "shape signature" of detected objects to an on-board database is an example of how
our work might be used for embedded change detection.
This work could be also extended by adding a polygon simplification scheme that takes advantage of properties 
of the Fourier decomposition, such as ease of differentiation.

\section*{Acknowledgements}

This project is funded
by the CIAR project at IRT Saint Exupéry.
The authors are grateful to the OPAL infrastructure from 
Université Côte d'Azur for providing resources and support.
This work was granted access to 
the HPC resources of IDRIS under the allocation 2020-
AD011011311R2 made by GENCI.
We thank Maxime Nabon for running our network on Jetson Xavier NX.

\appendix

\section{Chebyshev coefficients}

We demonstrate why Chebyshev coefficients, as used in \cite{Xu2019}, are not well 
suited to the regression of continuous periodic functions through a neural network.

Chebyshev polynomials are obtained through the following recurrence relation:

\begin{equation}
\begin{split}
         T_0(x) & = 1  \\
         T_1(x) & = x  \\
         T_{n+1}(x) & = 2 x\,T_n(x) - T_{n-1}(x). 
\end{split}
\end{equation}

Let $N$ be the number of degrees used for interpolation of $\rho$ using the 
truncated Chebyshev series. Then for $x \in [-1,1]$ we have:

\begin{equation}
    \displaystyle \rho (x) = \sum_{n=0}^{N} \alpha_{n} T_n(x)
\end{equation}

Since the contour $\rho$ is a continuous periodic function on $[-1,1]$, in order to perfectly
interpolate it, the Chebyshev coefficients $\alpha_i$ must satisfy:

\begin{equation}
    \displaystyle \sum_{n=0}^{N} \alpha_n T_n(-1) = \sum_{n=0}^{N} \alpha_n T_n(1)
\end{equation}

We also know that Chebyshev polynomials satisfy the following properties, for all $n \in \mathbb{N}$:

\begin{equation}
\begin{split}
    T_n(1) &= 1 \\
    T_{2n}(-1) &= 1 \\
    T_{2n+1}(-1) &= -1
\end{split}
\end{equation}

Then, we have the following:

\begin{equation}
    \displaystyle \sum_{k=0}^{N/2} \alpha_{2k} - \sum_{k=0}^{N/2} \alpha_{2k+1} = \sum_{n=0}^{N} \alpha_n
\end{equation}

Thus, we have shown that in order to perfectly interpolate a continuous periodic function, the Chebyshev coefficients must satisfy the following relationship:

\begin{equation}
    \displaystyle \sum_{k=0}^{N/2} \alpha_{2k+1} = 0 
\end{equation}

\begin{figure}[h!]
    \centering
    \includegraphics[width=\columnwidth]{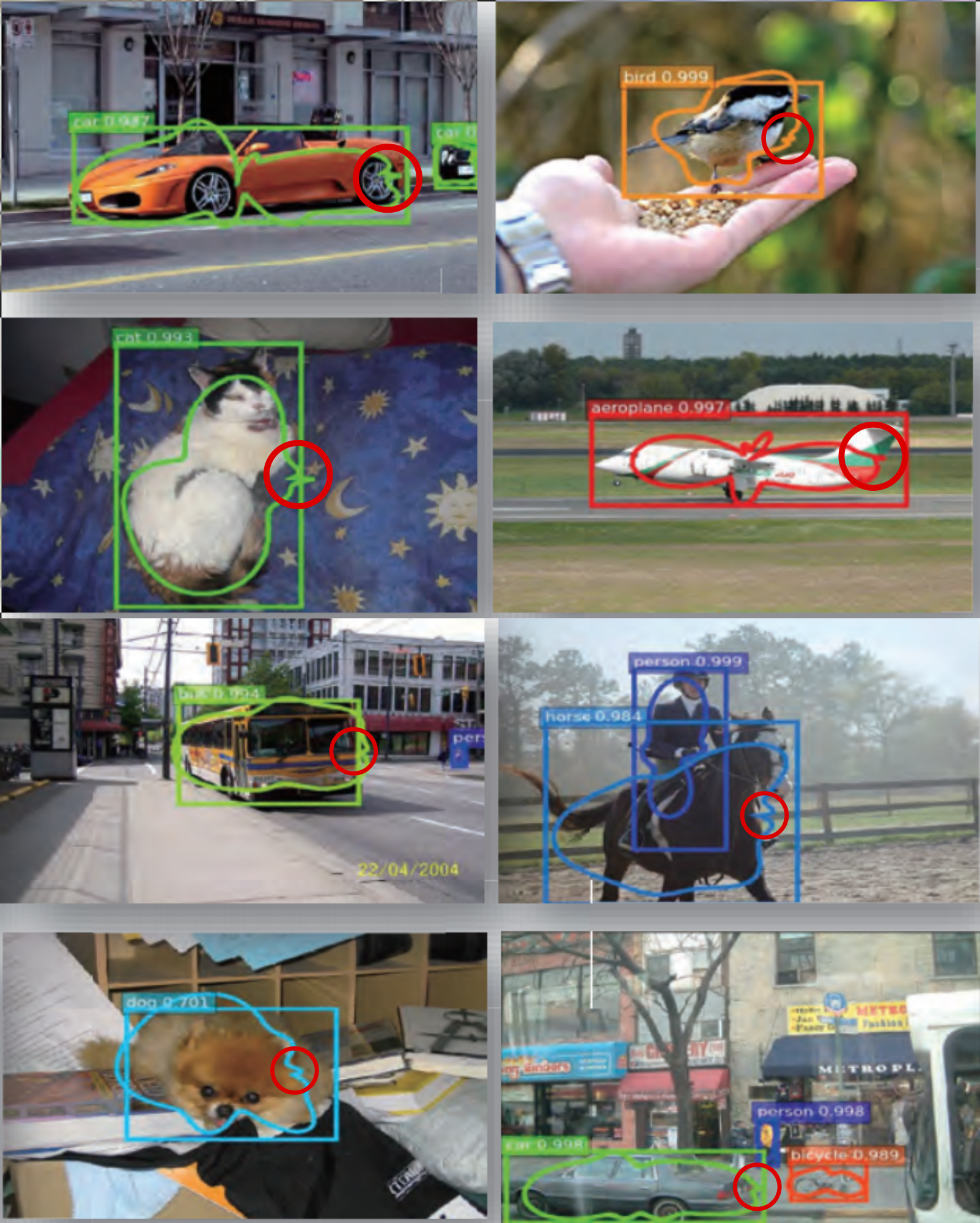}
    \caption{Discontinuities found in the results of \cite{Xu2019} when using Chebyshev coefficients for shape regression, circled in red.}
    \label{fig:discon}
\end{figure}

Figure \ref{fig:discon} shows the discontinuities created when this relationship
is not strictly enforced, e.g. when the Chebyshev coefficients are regressed by a neural network. 
They appear at $\theta = 2\pi$ on every regressed shape.

\section{Implementation details}

We use the MMDetection 2.7.0 object detection toolbox \cite{mmdetection}, based
on MMCV 1.2.1 for easy and fair comparison with other methods.
The underlying framework is PyTorch 1.7.0 with CUDA 10.1. We make use of 
the updated FFT package from PyTorch 1.7.0 which allows the use of complex numbers.
We use the Chamfer Distance implementation from PyTorch3D 0.3.0 \cite{ravi2020pytorch3d}. 
The interpolation scheme described in section 3 of the main text is implemented as a module in the MMDetection
data loading pipeline using the \texttt{interp} module from Numpy.

\begin{table}[h!]
    \centering
\setlength\tabcolsep{2pt}
    \begin{tabular}{c|c|c|c|c|c}
         Name-Backbone & DCNv2 & CARAFE & $N_{in}$ & $N_{feat}$ & Time  \\ \hline
         SCR-D53-320 & No & No & 128 & 128 & 20h \\
         SCR-R50-800 & Yes & Yes & 256 & 256 & 40h \\
         SCR-X101-800 & Yes & Yes & 256 & 256 & 80h \\
    \end{tabular}
    \caption{Characteristics of our main networks. Variations are further explored in our ablation study.
    DCN refers to the use of Deformable Convolution v2 \cite{Zhu} layers in the head. CARAFE refers to the use of the
    FPN-CARAFE \cite{Wangc} neck. Training time on 8 Tesla V100 GPUs.}
    \label{tab:hyperp}
\end{table}

Our networks are trained for 38 epochs using SGD with momentum (0.9), with an initial learning
rate of 0.0003, batch-size of 16 (2 images per GPU), 
gradient clipping at a maximum $L^2$ norm of 45, and employ of a 500 step warm-up with a ratio of 0.001 applied to the learning rate.
We use the multiscale training feature found in MMDetection to train with image heights of 640 and 800 pixels.
We balanced our loss function terms through trial-and-error, and settled on the following values: 
$\lambdaloss{cls}$=1.0, $\lambdaloss{cent}$=1.0, $\lambdaloss{CD}$=1.0, $\lambdaloss{perim}$=0.01, $\lambdaloss{coeff}$=500.0.

Due to software incompatibilities during the conversion of our DarkNet-53 model for low-power hardware targets,
we had to simplify the architecture by removing deformable convolutions and CARAFE, as shown in Table \ref{tab:hyperp}.

\section{Hardware configuration}

This section contains details regarding the embedded hardware configuration used for the throughput benchmarks presented in section 4 of the main text.

Our Nvidia 
Jetson Nano 2GB, Nano 4GB and Xavier NX development kits are running JetPack 4.5, 
ONNXRuntime 1.4.0 and TensorRT 7.1.3. 
The Nano SoC features a quad-core ARM A53 CPU and
a 128 CUDA core Maxwell GPU. In order to obtain consistent
results, we lock the power consumption at
10W for benchmarking using the \texttt{jetson\_clocks} tool and cool
the heatsink with a Noctua A4x20 5V PWM fan using the default fan curve.
We also ran our benchmarks on a Jetson Nano 4GB model and found
no performance difference with the 2 GB model. Thus, Table 6 of the main text 
refers to both of them as "Nano".

The Jetson Xavier NX features a 6-core ARM processor, a 384 CUDA core
Volta GPU and 8GB RAM, with a power consumption of 30W and is running the same setup.

The Intel Movidius Myriad X Vision Processing Unit is an ultra low-power AI
accelerator with a power consumption of only 1W. It has been used in ESA's $\phi$-sat-1 CubeSat,
as well as in smart security cameras,
UAVs and industrial machine vision equipment. We benchmark SCR
on the Myriad X using OpenVino and FP16 precision.

Nvidia GeForce GTX 1080Ti (11GB VRAM, 250W) benchmarks were run using either the same MMDetection setup used for training,
or TensorRT 7.1.3.

Intel Xeon Silver 4114 CPU benchmarks were run using the Intel OpenVino toolkit.

\section{Carbon Impact Statement}

While the focus of our work is to create efficient neural networks 
that are able to run on very low power devices, we cannot help but 
notice that these networks are still created and trained using
power-hungry multi-GPU machines and wonder about the environmental 
impact. For this paper, we missed the opportunity to track the real power consumption of our experiments, and to estimate its global environmental footprint, but we are trying as much as we can to publish these estimations, as recommended in \cite{henderson2020towards,anthony2020carbontracker,lannelongue2020green}.

In order to run the experiments required for our main results and 
ablation study, we have used 5864.55 GPU hours on Nvidia Tesla V100
GPUs, which are rated for a power consumption of 300W. This, not counting
CPUs, cooling, PSU efficiency, storage of datasets and results,
as well as different trials or hyper-parameter searches on workstations,
amounts to 1759kWh. Since the carbon intensity of our electricity grid is 10
gCO2/kWh, we estimate an emission of 17590 gCO2, which is equivalent to 146 km traveled by car according to \cite{anthony2020carbontracker}.

\vspace{3em}
\section{Qualitative results}

In this section, we provide an extended selection of qualitative results that could not be part of the main paper for space reasons.
Figures \ref{fig:ext_qual} and \ref{fig:ext_qual2} show results from our smallest and fastest DarkNet-53-based model.
 
\setlength{\teaserheight}{0.16\textwidth}
\setlength{\teaserheightt}{0.1617\textwidth}

\begin{figure*}[p!]
    \centering
\includegraphics[height=1.07\teaserheight]{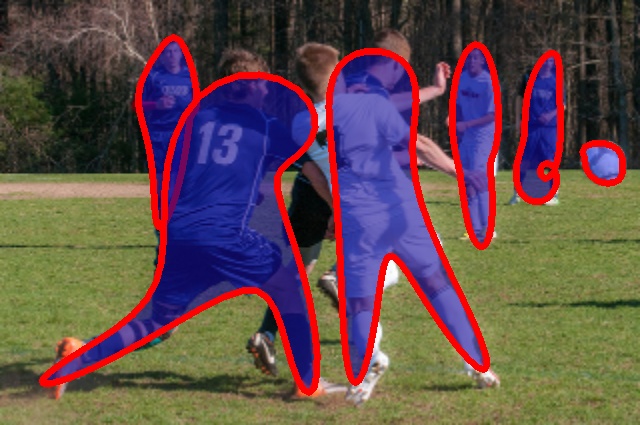} \hfill
\includegraphics[height=1.07\teaserheight]{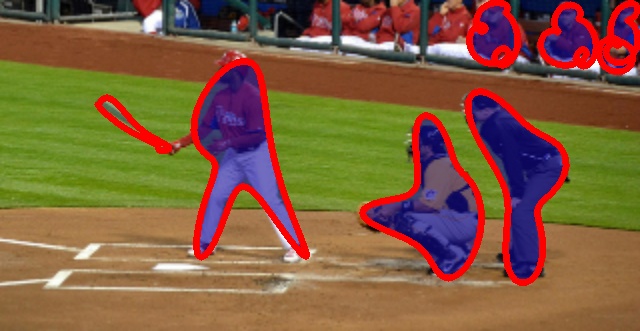} \hfill
\includegraphics[height=1.07\teaserheight]{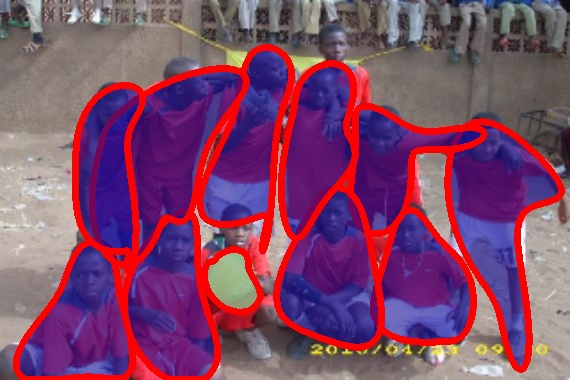} \hfill
\includegraphics[height=1.07\teaserheight]{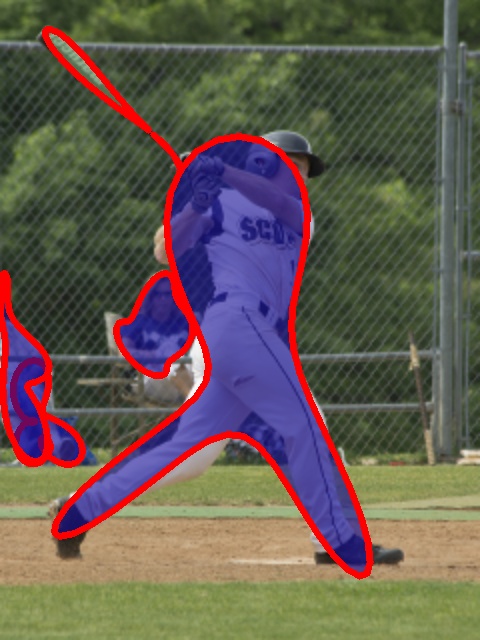}

\includegraphics[height=1.14\teaserheight]{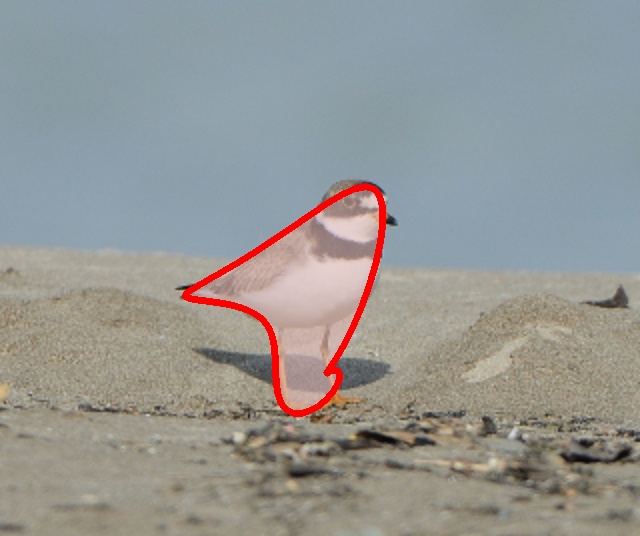} \hfill
\includegraphics[height=1.14\teaserheight]{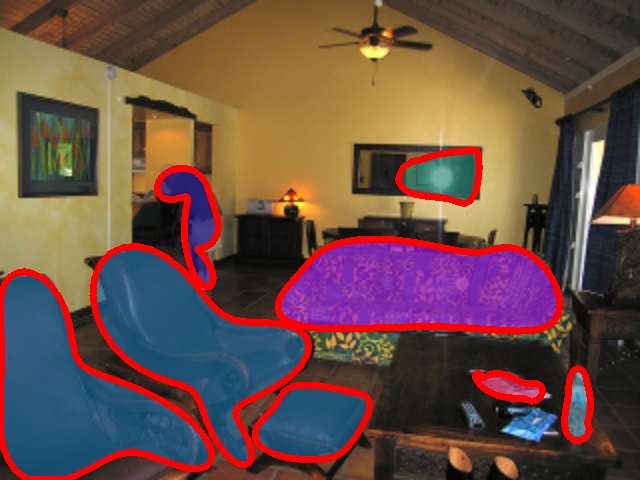} \hfill
\includegraphics[height=1.14\teaserheight]{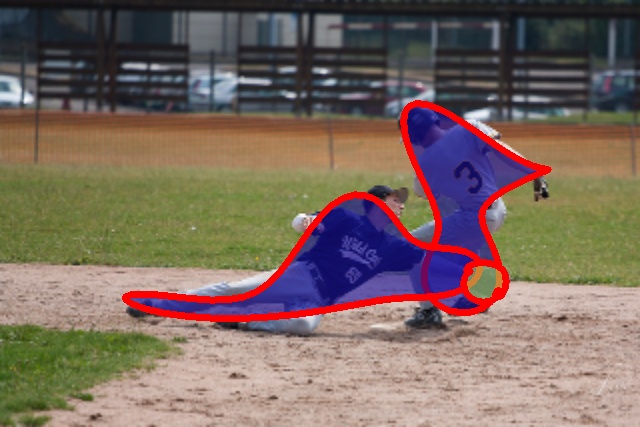} \hfill
\includegraphics[height=1.14\teaserheight]{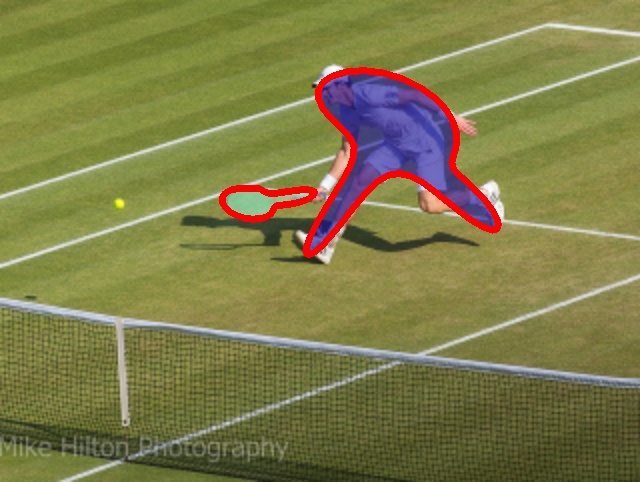}

\includegraphics[height=1.1\teaserheight]{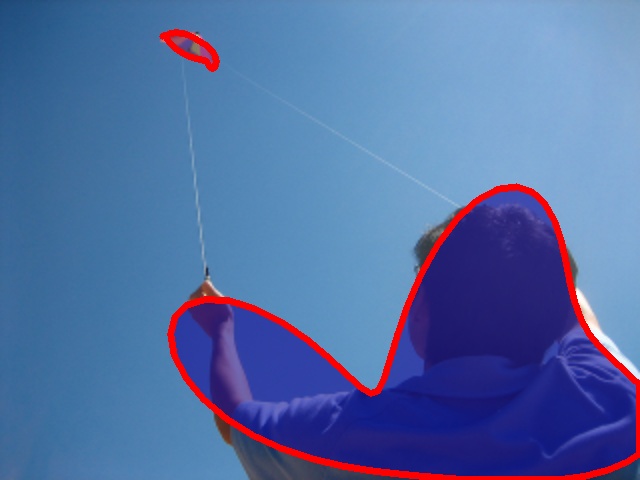} \hfill
\includegraphics[height=1.1\teaserheight]{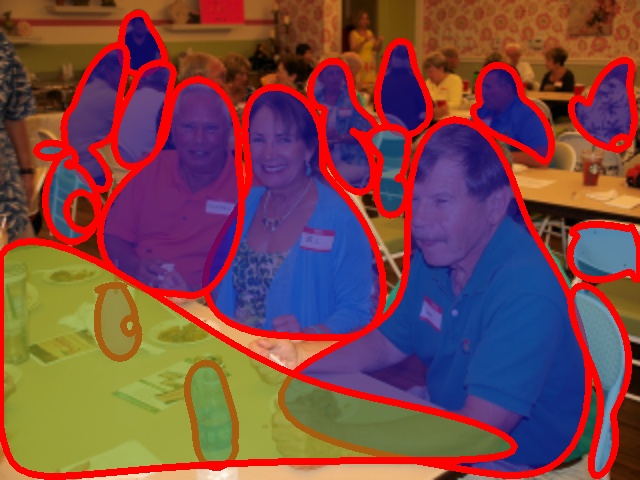} \hfill
\includegraphics[height=1.1\teaserheight]{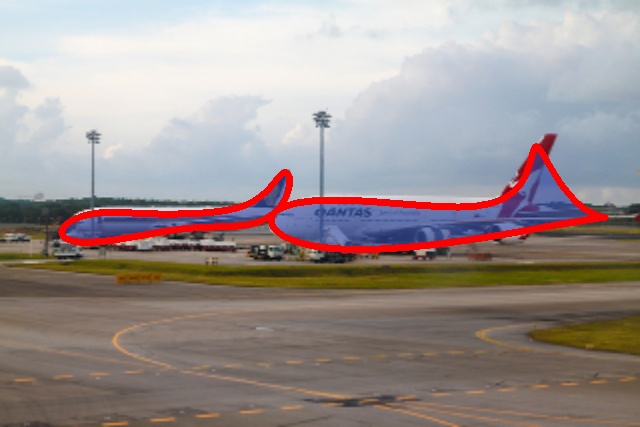} \hfill
\includegraphics[height=1.1\teaserheight]{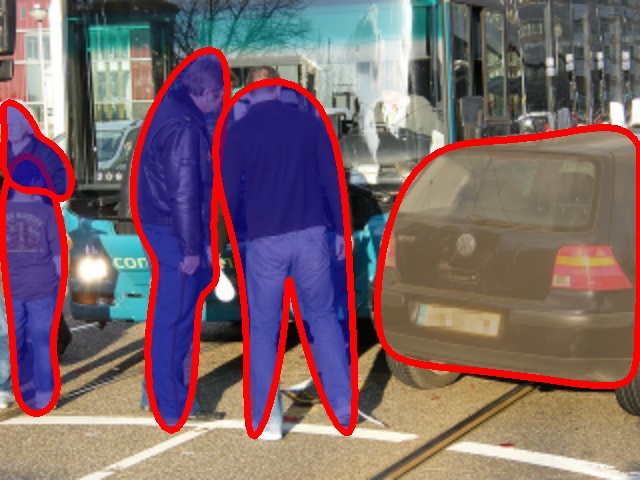}

\includegraphics[height=1.08\teaserheight]{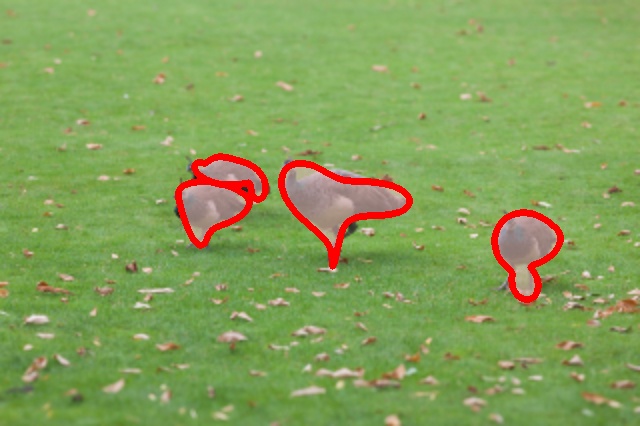} \hfill
\includegraphics[height=1.08\teaserheight]{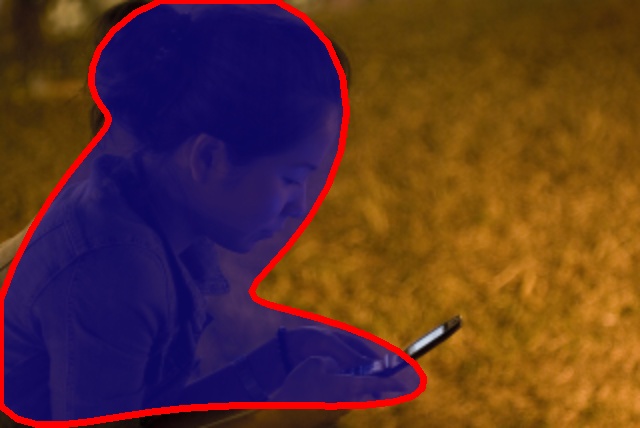} \hfill
\includegraphics[height=1.08\teaserheight]{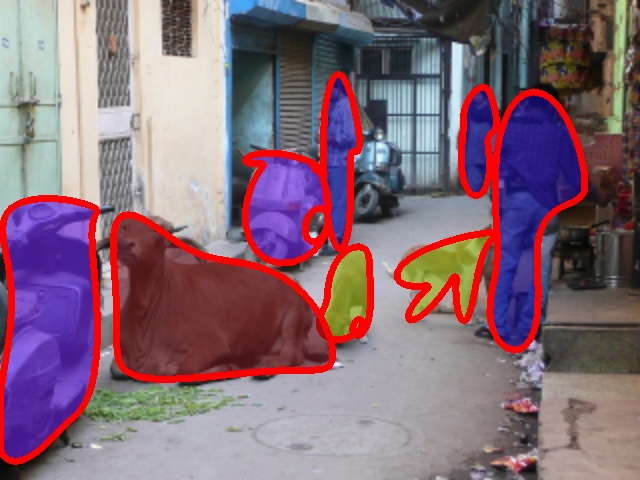} \hfill
\includegraphics[height=1.08\teaserheight]{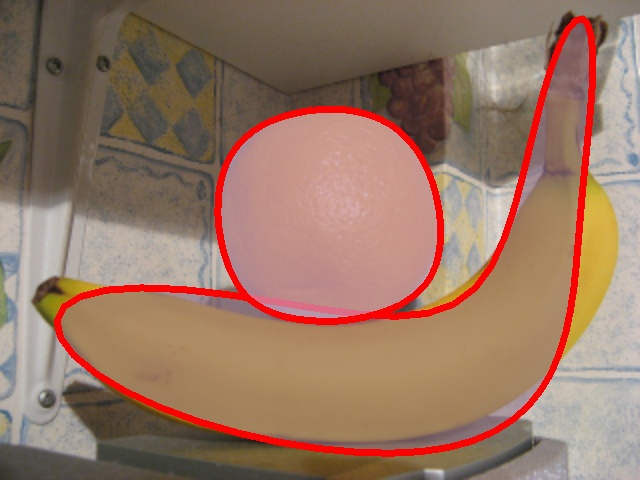}

\setlength{\teaserheight}{0.19\textwidth}
\includegraphics[height=\teaserheight]{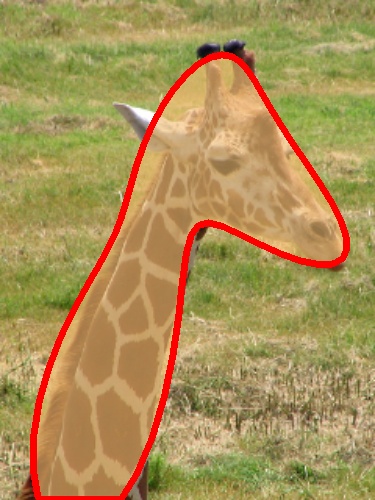} \hfill
\includegraphics[height=\teaserheight]{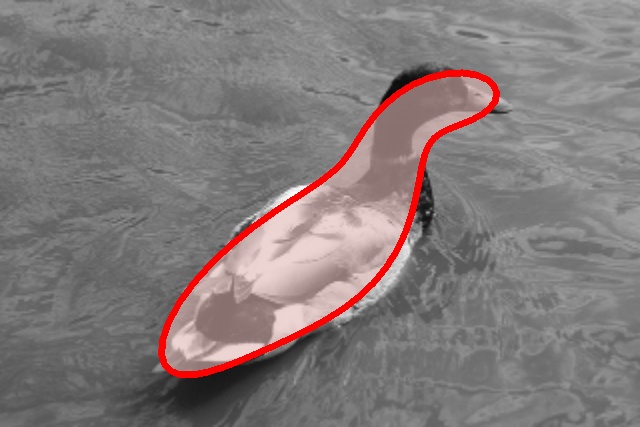} \hfill
\includegraphics[height=\teaserheight]{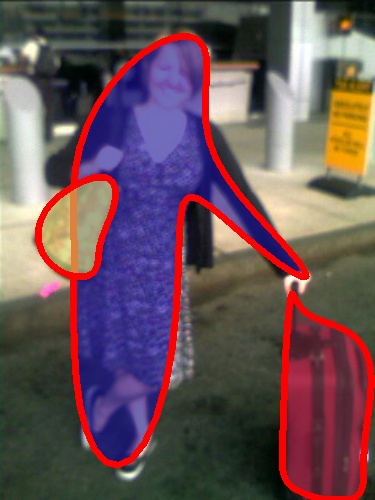} \hfill
\includegraphics[height=\teaserheight]{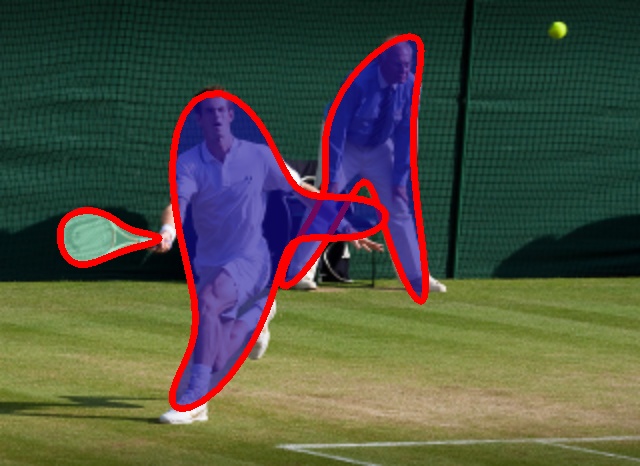} \hfill
\includegraphics[height=\teaserheight]{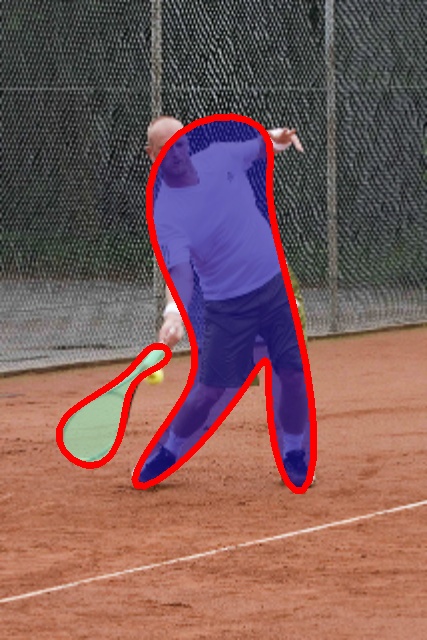}

\setlength{\teaserheight}{0.16\textwidth}
\includegraphics[height=1.15\teaserheight]{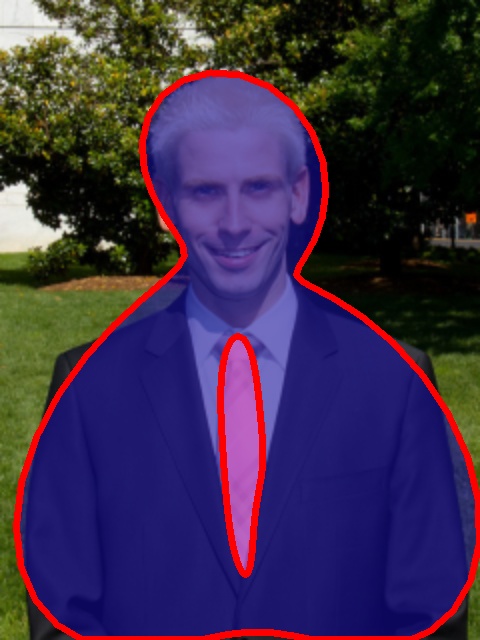} \hfill
\includegraphics[height=1.15\teaserheight]{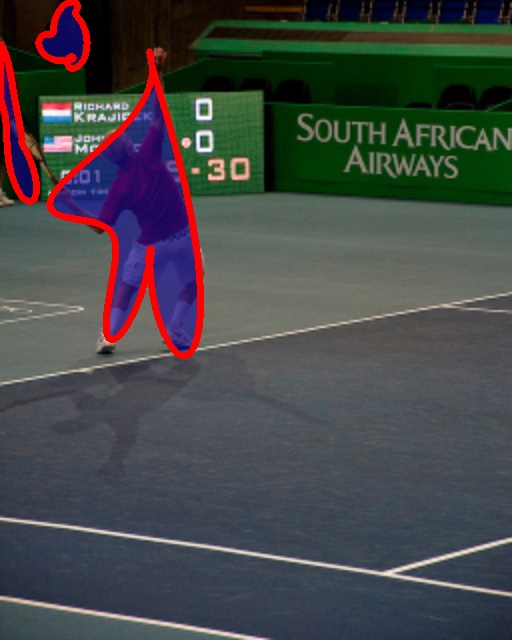} \hfill
\includegraphics[height=1.15\teaserheight]{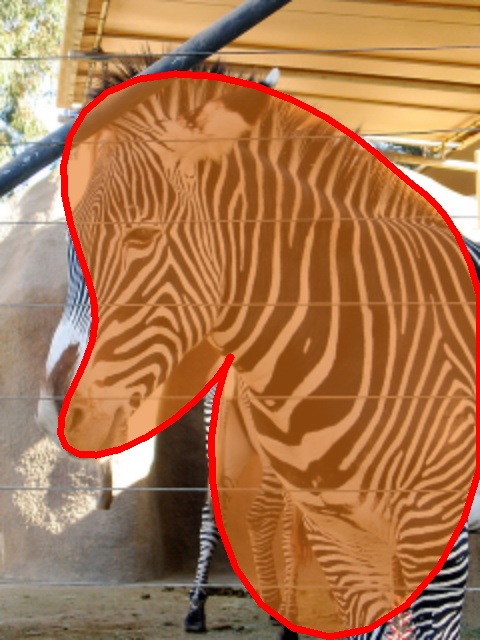} \hfill
\includegraphics[height=1.15\teaserheight]{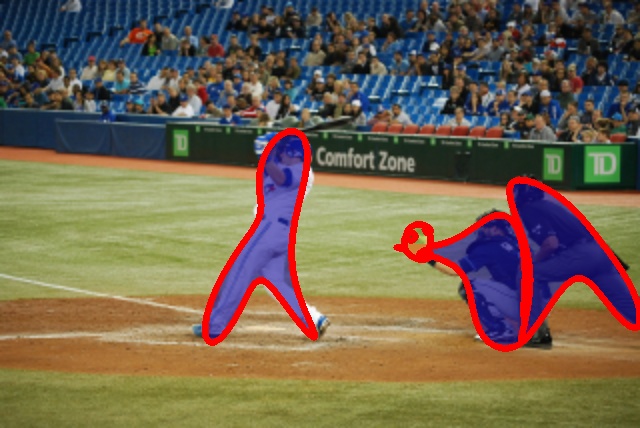} \hfill
\includegraphics[height=1.15\teaserheight]{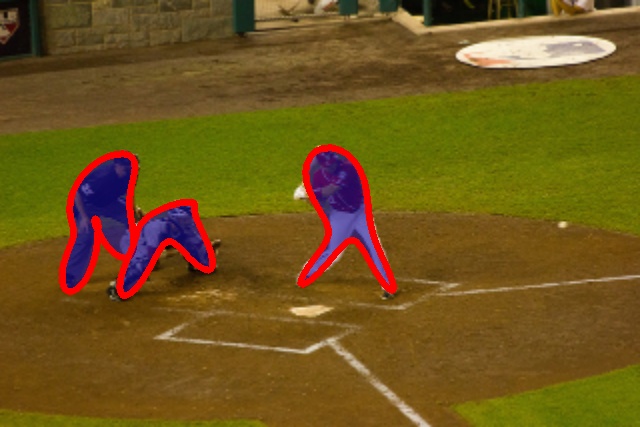}

    \caption{More results from our DarkNet-53 based model selected from COCO 2017 test-dev.}
    \label{fig:ext_qual2}
\end{figure*}

\setlength{\teaserheight}{0.16\textwidth}

\begin{figure*}[p!]
    \centering
\includegraphics[height=0.19\textwidth]{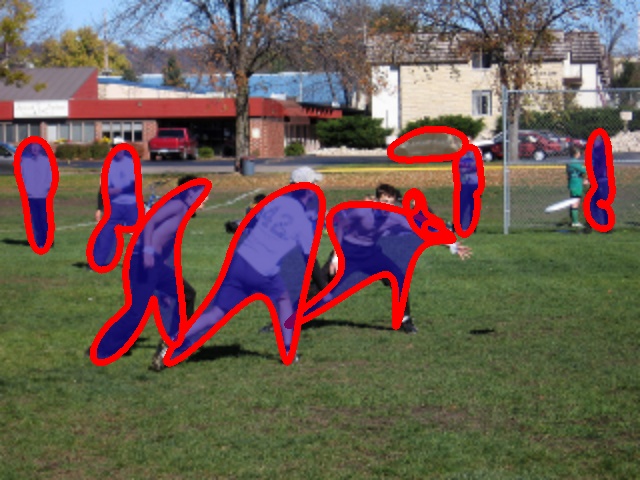}  \hfill
\includegraphics[height=0.19\textwidth]{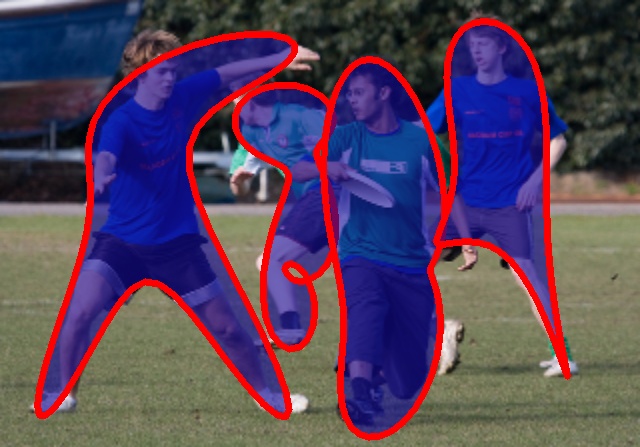} \hfill
\includegraphics[height=0.19\textwidth]{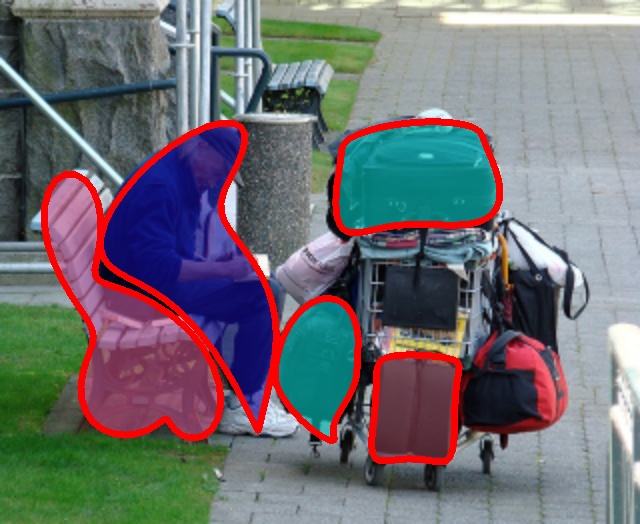} \hfill
\includegraphics[height=0.19\textwidth]{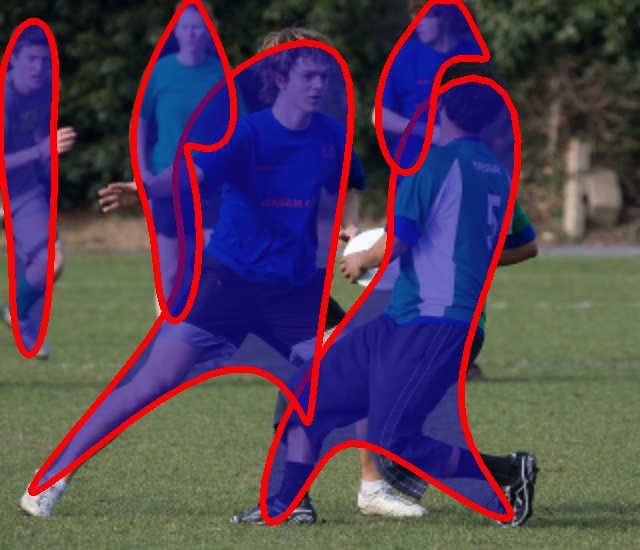}

\includegraphics[height=1.05\teaserheight]{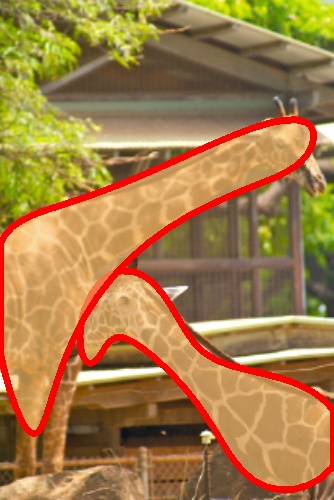} \hfill
\includegraphics[height=1.05\teaserheight]{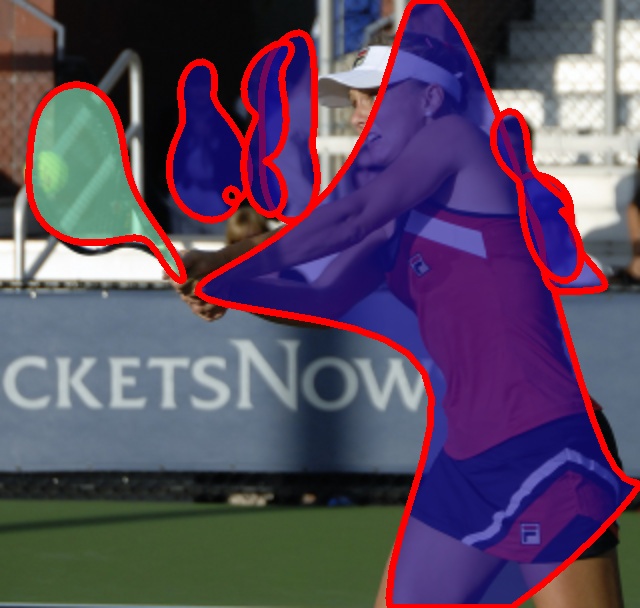} \hfill
\includegraphics[height=1.05\teaserheight]{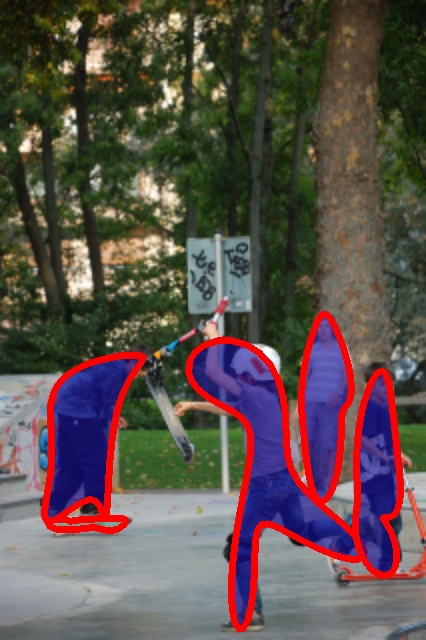} \hfill
\includegraphics[height=1.05\teaserheight]{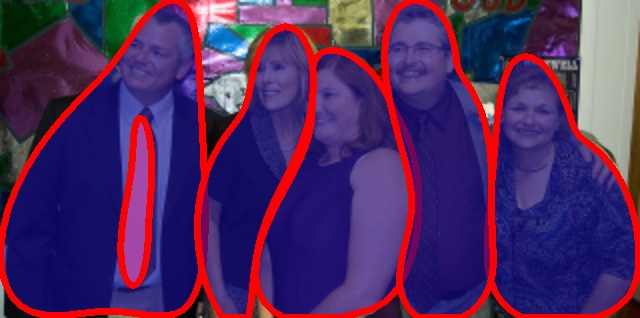} \hfill
\includegraphics[height=1.05\teaserheight]{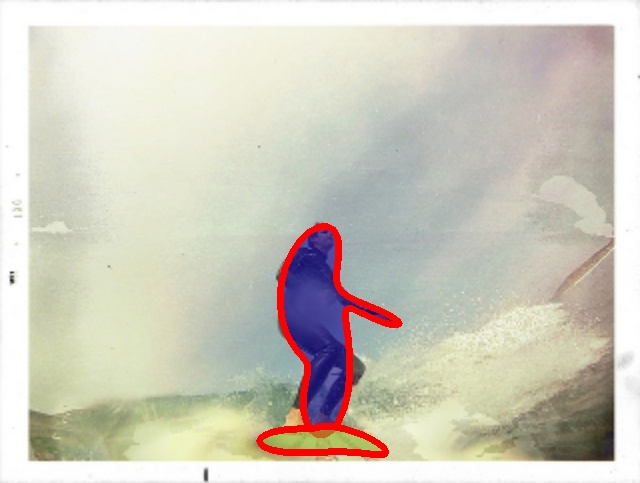}

\includegraphics[height=0.172\textwidth]{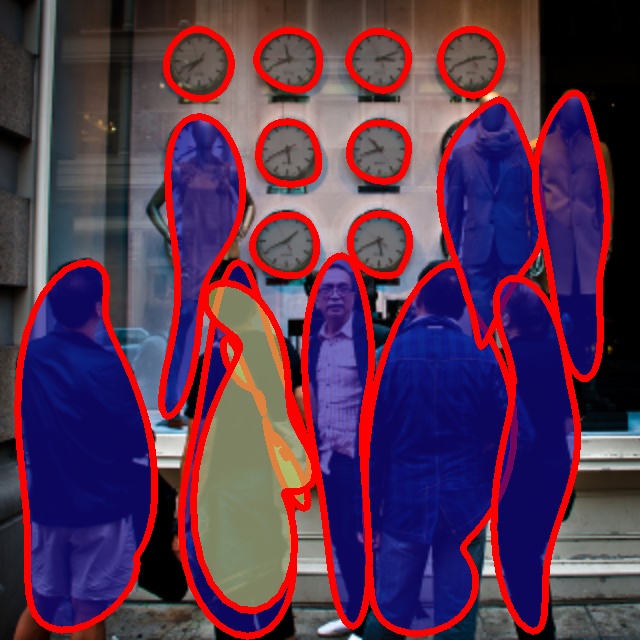} \hfill
\includegraphics[height=0.172\textwidth]{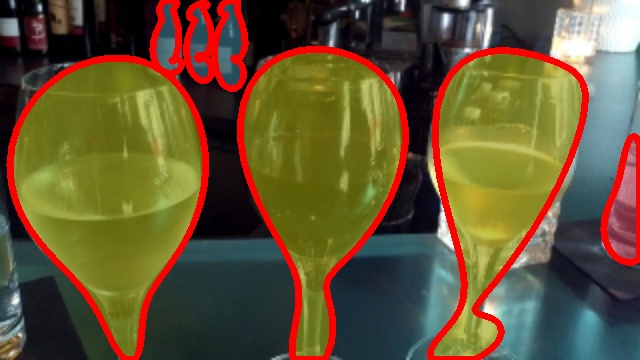} \hfill
\includegraphics[height=0.172\textwidth]{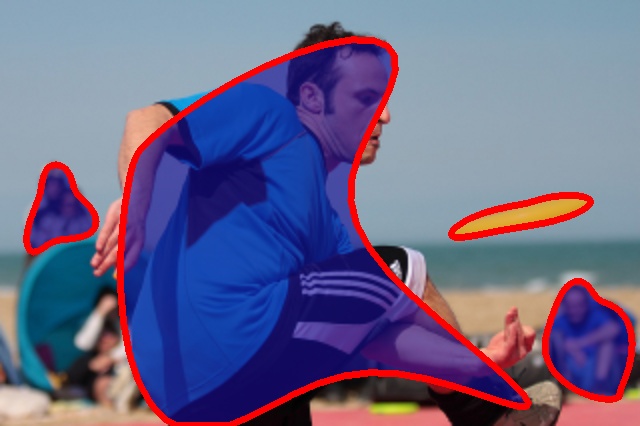} \hfill
\includegraphics[height=0.175\textwidth]{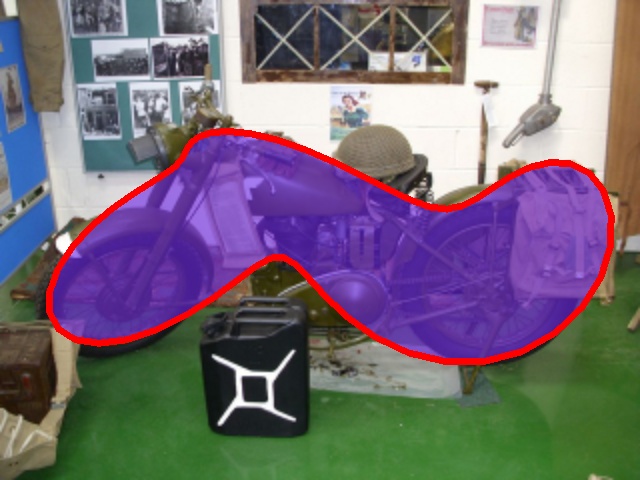}

\includegraphics[height=1.02\teaserheight]{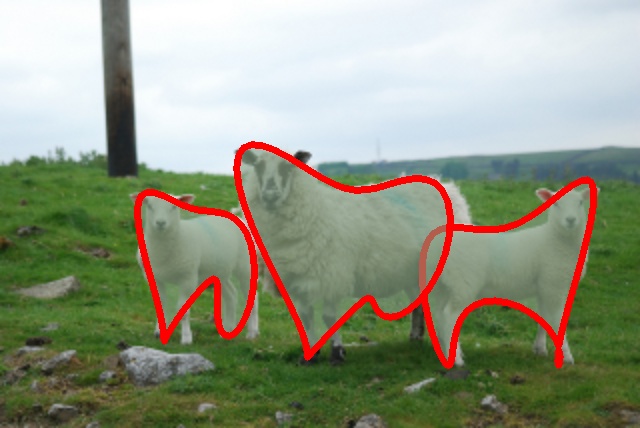} \hfill
\includegraphics[height=1.02\teaserheight]{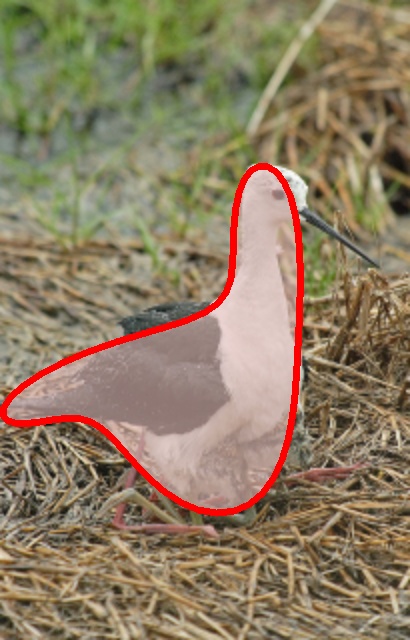} \hfill
\includegraphics[height=1.02\teaserheight]{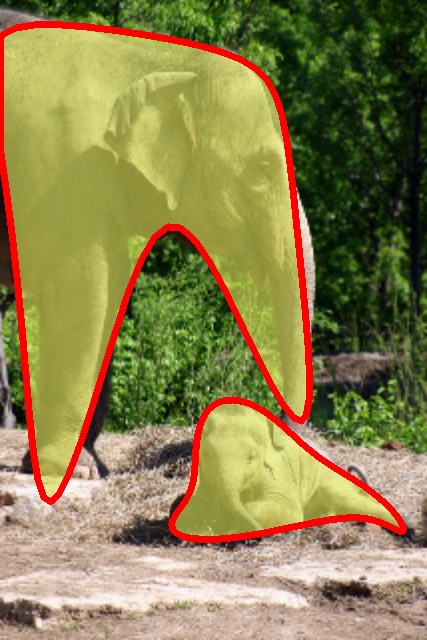} \hfill
\includegraphics[height=1.02\teaserheight]{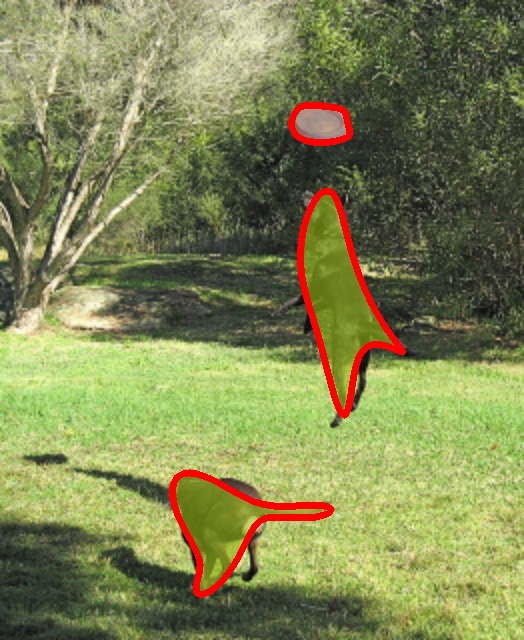} \hfill
\includegraphics[height=1.02\teaserheight]{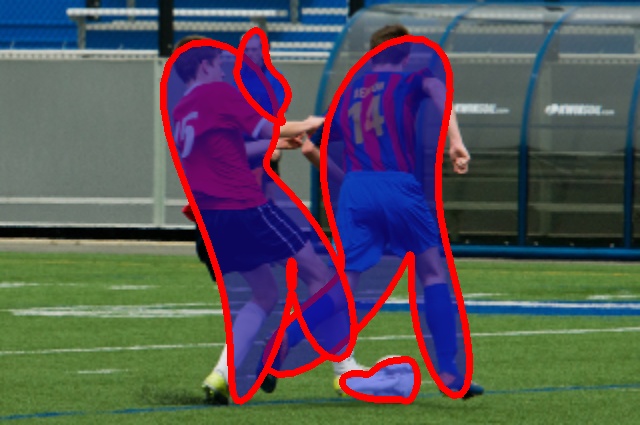} \hfill
\includegraphics[height=1.02\teaserheight]{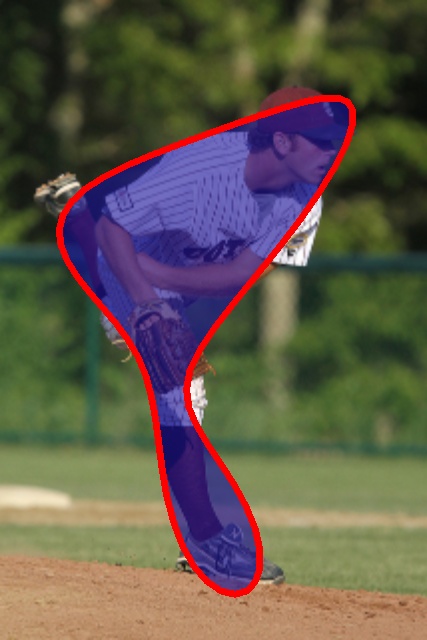}

\includegraphics[height=1.03\teaserheight]{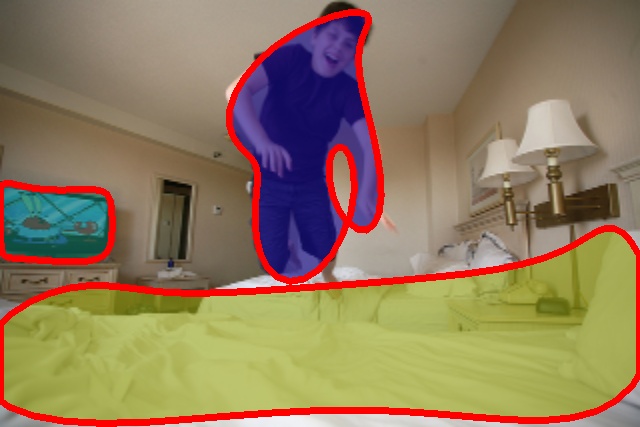} \hfill
\includegraphics[height=1.03\teaserheight]{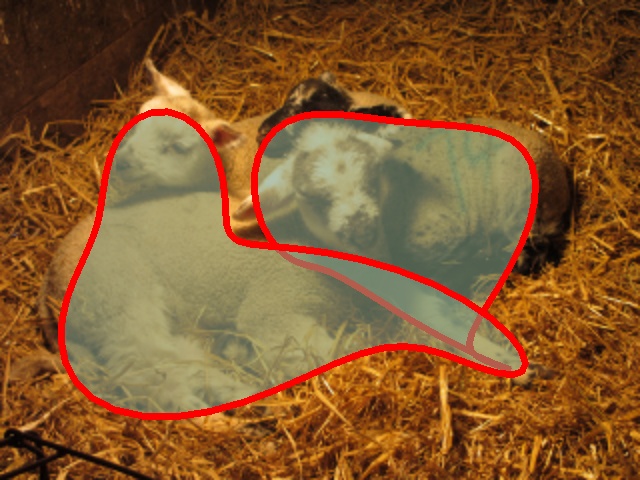} \hfill
\includegraphics[height=1.03\teaserheight]{figures/selec_0.3/000000111483.jpg} \hfill
\includegraphics[height=1.03\teaserheight]{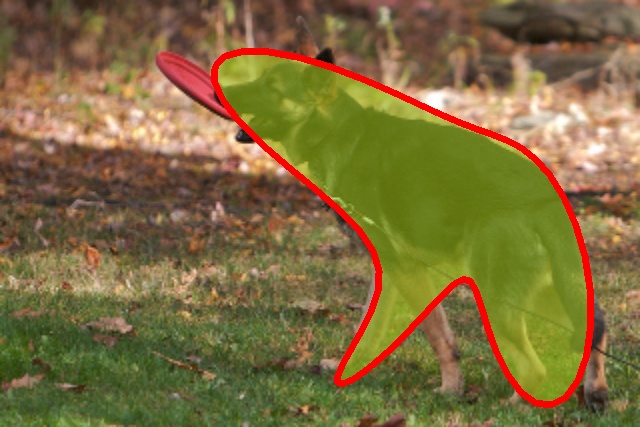}

\includegraphics[height=1.02\teaserheight]{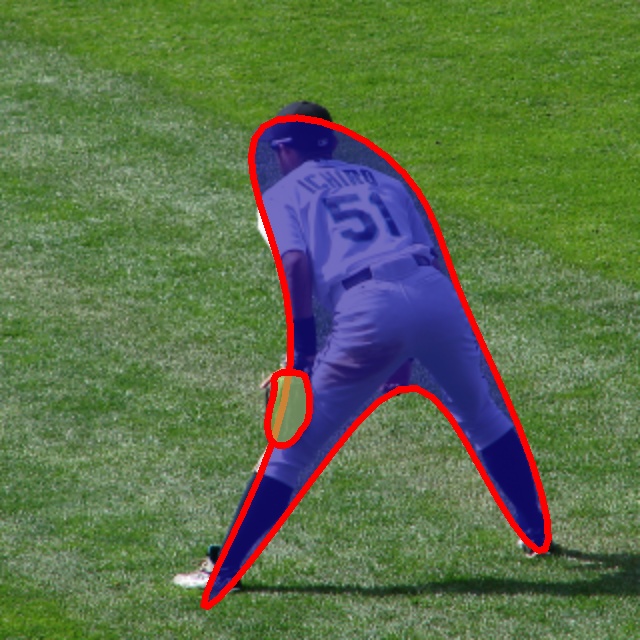} \hfill
\includegraphics[height=1.02\teaserheight]{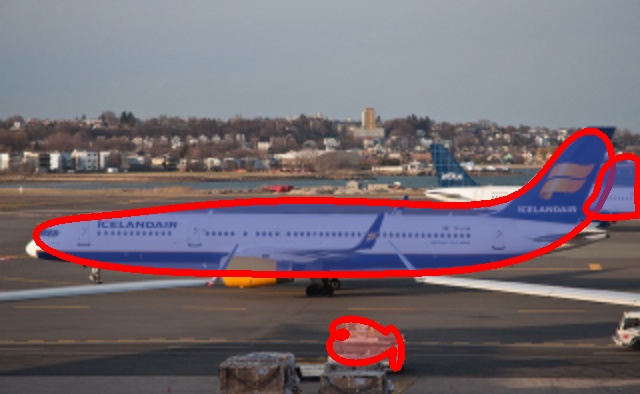} \hfill
\includegraphics[height=1.02\teaserheight]{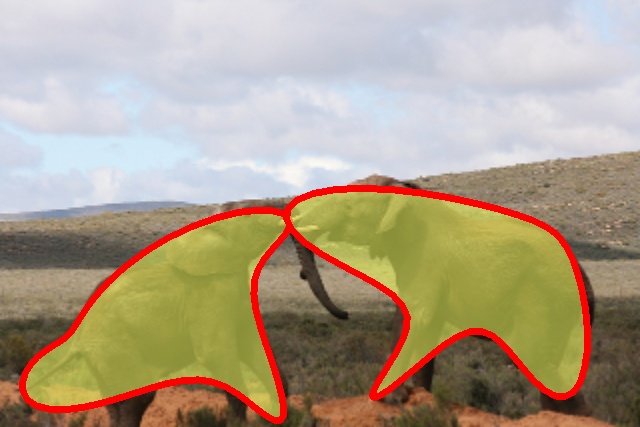} \hfill
\includegraphics[height=1.02\teaserheight]{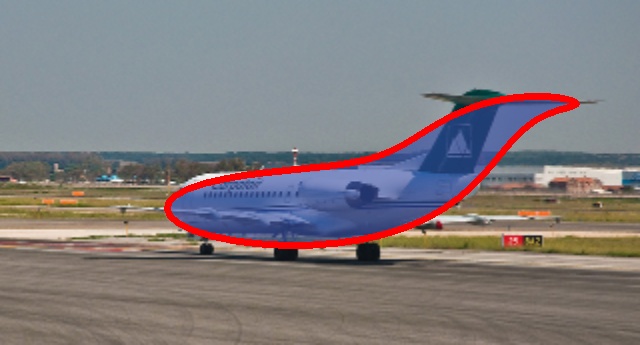}

\includegraphics[height=1.05\teaserheight]{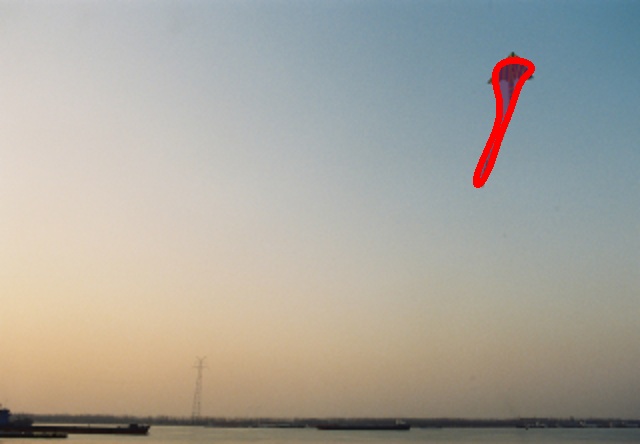} \hfill
\includegraphics[height=1.05\teaserheight]{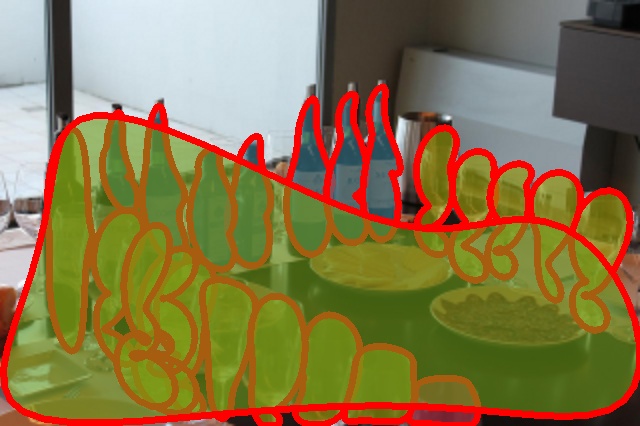} \hfill
\includegraphics[height=1.05\teaserheight]{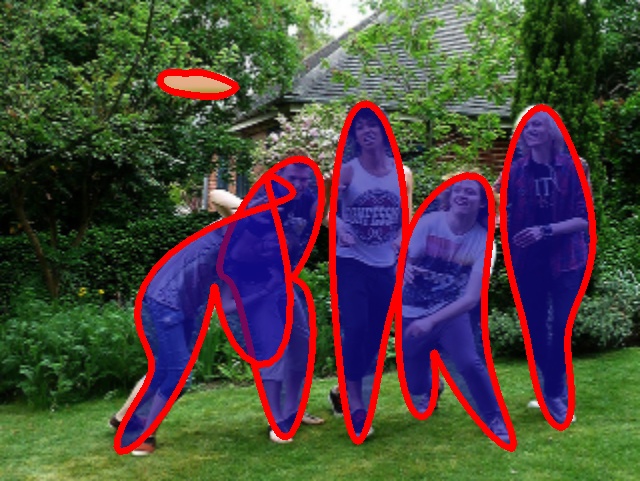} \hfill
\includegraphics[height=1.05\teaserheight]{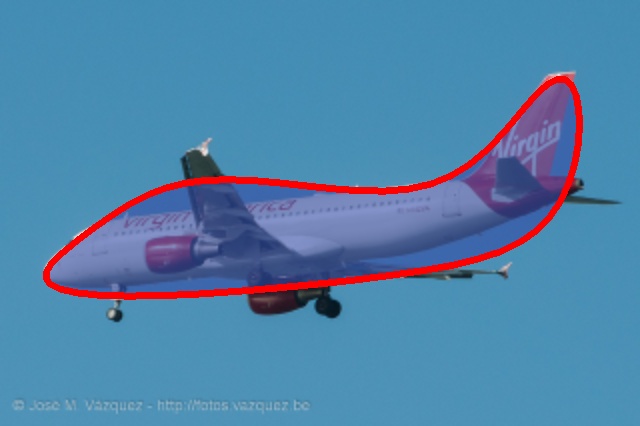}

    \caption{More results from our DarkNet-53 based model selected from COCO 2017 test-dev.}
    \label{fig:ext_qual}
\end{figure*}

\clearpage

{\small
\bibliographystyle{ieee_fullname}
\bibliography{bib_contour_clean}
}

\end{document}